\documentclass[runningheads]{llncs}
\usepackage{amssymb,amsmath}
\usepackage{graphicx}
\usepackage[strings]{underscore}
\usepackage{longtable}
\usepackage{booktabs}
\usepackage[load-configurations=version-1]{siunitx}
\usepackage{appendix}
\usepackage{todonotes}
\usepackage{semantic}
\usepackage{multirow}
\usepackage{array}
\usepackage{algpseudocode}
\usepackage{algorithm}
\usepackage{hyperref}
\usepackage{caption}
\usepackage{subcaption}
\usepackage{hhline}
\usepackage{changepage}
\usepackage{adjustbox}
\usepackage{theorem}
\usepackage{cleveref}

\theorembodyfont{\upshape}
\theoremheaderfont{\scshape}

\newcommand{\Sent}{\mathcal{S}_{\models}}
\newcommand{\Scon}{\mathcal{S}_{\bot}}
\newcommand{\Sund}{\mathcal{S}_{?}}
\newcommand{\Mod}{\mathit{Mod}}
\begin{document}
\title{A hierarchy of faithfulness criteria for
knowledge base completion}
\titlerunning{Faithfulness in KBC}
% If the paper title is too long for the running head, you can set
% an abbreviated paper title here
%
\author{Olga Mashkova\inst{1}\orcidID{0000-0002-4916-1660} and Robert Hoehndorf\inst{1}\orcidID{0000-0001-8149-5890}}
\authorrunning{Mashkova and Hoehndorf}
% First names are abbreviated in the running head.
% If there are more than two authors, 'et al.' is used.
%
\institute{Computer Science Program, Computer, Electrical, and
  Mathematical Sciences \& Engineering Division, King Abdullah
  University of Science and Technology, Thuwal 23955, Saudi Arabia\\
\email{\{first\_name.last\_name\}@kaust.edu.sa}}
\maketitle              % typeset the header of the contribution
\begin{abstract}
Knowledge graph completion is evaluated by ranking observed triples
above randomly corrupted ones, which treats every unobserved fact as
false. When the object being completed is a description logic knowledge
base rather than a plain graph, the open world assumption and deductive
closure make this inadequate: relative to the knowledge base, a
candidate axiom is entailed, contradictory, or undetermined, and a model
that cannot separate a logically impossible axiom from a plausible novel
one is not merely less accurate but semantically incorrect. We ask what it
means for a knowledge base completion model to be logically faithful,
and whether current embedding models are. We define a hierarchy of four
increasingly strict criteria, discrimination, logical admissibility,
monotonic logical faithfulness, and probabilistic logical faithfulness,
and prove that they form a strict chain of implications. We ground the
strongest criterion in the relative model count
$P(\alpha\mid\mathcal{O}) = \#(\mathcal{O}\cup\{\alpha\})/\#(\mathcal{O})$,
which recovers the trichotomy at its endpoints and ranks undetermined
axioms in between. Evaluating knowledge graph and logic-geometric
embedding models on $\mathcal{EL}$ ontologies, with entailed, contradictory, and
undetermined test sets generated by a reasoner, we find that ranking
accuracy does not imply logical faithfulness and that none of the
evaluated models is faithful across the hierarchy. The code is available at \url{https://github.com/bio-ontology-research-group/kbc}.
\end{abstract}

\section{Introduction}

Knowledge graph completion (KGC) is formulated as the prediction of
missing facts in the form of $(head, relation, tail)$ triples,
identifying new edges between entities within a graph
structure~\cite{chen2020knowledge}. Standard metrics such as Hits@$n$ and mean reciprocal rank (MRR) rank
observed test triples against corrupted or otherwise unobserved
ones~\cite{bordes2013translating}, and the dominant training objective
(random negative sampling) treats unobserved triples uniformly as
negatives. Both the training signal and the evaluation protocol thus
implicitly identify ``not in the KG'' with ``false''.

The transition from knowledge graphs to knowledge bases (KBC)
introduces phenomena that this collapse cannot accommodate. Knowledge
bases (or ontologies) include terminological axioms (TBox), such as
concept subsumption $C \sqsubseteq D$, equivalence $C \equiv D$, and
disjointness $C \sqcap D \sqsubseteq \bot$, and under the open world
assumption (OWA) the absence of an axiom does not imply that it is
false. The \emph{entailment status} of a candidate axiom $\alpha$
relative to a knowledge base $\mathcal{O}$ is therefore not binary but
a trichotomy: $\alpha$ may be \emph{entailed} (provably true),
$\mathcal{O} \models \alpha$; \emph{contradictory} (provably false),
$\mathcal{O} \models \neg\alpha$, equivalently
$\mathcal{O}\cup\{\alpha\} \models \bot$; or \emph{undetermined}
(logically possible), neither proven nor disproven and hence a valid
potential extension of the KB.

A KBC model whose scores conflate \emph{contradictory} with
\emph{undetermined} axioms is not merely suboptimal: it is unable to
distinguish a logical impossibility from a plausible novel
prediction. By training to minimise the score of all unobserved triples
via random negative sampling, mainstream KGC methods force statements
in $\Sund$, which are exactly the candidate \emph{discoveries} a KBC
system is meant to surface, into the same low-scoring region as
statements in $\Scon$. This is not just a calibration problem in the
sense of~\cite{tabacof2020probability}: any downstream consumer that
uses scores to filter candidates will accept logically impossible
axioms as often as plausible ones in expectation.

We address this gap by introducing a four-level \emph{hierarchy of
faithfulness criteria} that any KB completion model can be evaluated
against. From weakest to strongest: \emph{discrimination} (L0) requires
every entailed axiom to outscore every contradiction, an averaged
surrogate of which classical KGC metrics measure; \emph{logical
admissibility} (L1) requires that no contradiction outscore any
consistent axiom; \emph{monotonic logical faithfulness} (MLF, L2)
requires the strict trichotomy ordering $f(\alpha^{-}) < f(\alpha^{?}) <
f(\alpha^{+})$; and \emph{probabilistic logical faithfulness} (PLF, L3)
requires $f$ to be monotonic in the relative model count
$P(\alpha\mid\mathcal{O}) =
\#(\mathcal{O}\cup\{\alpha\})/\#(\mathcal{O})$, so that ranking within
$\Sund$ matches the proportion of models in which $\alpha$ holds
(Section~\ref{sec:hierarchy} gives formal definitions).

Our contributions are as follows. We formalise the KBC task as distinct
from KGC under deductive closure and OWA
(Section~\ref{sec:prelim}), make the entailment trichotomy precise, and
show that the relative model count $P(\alpha\mid\mathcal{O})$ recovers
it at its endpoints ($P=0$ on $\Scon$, $P=1$ on $\Sent$) and orders
$\Sund$ in between (Section~\ref{sec:foundations}). We define the
four-level hierarchy and prove the strict implication chain
$\mathrm{PLF}\Rightarrow\mathrm{MLF}\Rightarrow\text{admissibility}
\Rightarrow\text{discrimination}$
(Section~\ref{sec:hierarchy}, Theorem~\ref{thm:hierarchy}), and identify
training-time interventions that target the faithfulness gap
(Section~\ref{sec:training}). Finally, we evaluate seven geometric and
logic-geometric models against the hierarchy on $\mathcal{EL}$
ontologies, with reasoner-generated test sets, and find that ranking
accuracy does not imply faithfulness, that a model's position on the
hierarchy depends on the logical construct, and, via a
negative-sampling sweep, that substituting contradiction-based for
random negatives improves faithfulness at a cost in ranking accuracy
(Section~\ref{sec:experiments}).

\section{Related work}
\label{sec:related}

Embedding-based KGC, including TransE~\cite{bordes2013translating} and
the factorisation family (DistMult~\cite{yang2015embedding},
ComplEx~\cite{trouillon2016complex},
RESCAL~\cite{nickel2011three}), relies on graph topology and ranks
observed triples against random corruptions. That this misbehaves under
the open world assumption has been argued
semantically~\cite{ceylan2021open} and for the standard ranking
metrics~\cite{yang2022rethinking}; removing entailed triples that leak
from train into test through deductive closure shows reported scores to
be inflated~\cite{akrami2020realistic}; surveys catalogue the gap
between predicted scores and logical
admissibility~\cite{liang2024survey,peng2023knowledge}; and
region-based analyses show that common embedding geometries cannot
capture even simple rule sets~\cite{gutierrez2018knowledge}. Closer to
our goal, consistency-aware methods filter predictions that violate the
ontology~\cite{jain2021improving}, and semantic-aware metrics such as
Sem@K score type-constraint satisfaction in the top
ranks~\cite{hubert2023semk}; both check a necessary admissibility
condition, but neither isolates entailed axioms nor ranks the
undetermined region. Our hierarchy makes the gap quantitative: current
models partially satisfy discrimination (L0) but fail MLF (L2) by
construction, because the loss gives no incentive to distinguish
$\Scon$ from $\Sund$.

A separate line of work explicitly encodes the geometry of subsumption
and disjointness: ELEmbeddings~\cite{kulmanov2019embeddings} maps
$\mathcal{EL}$ concepts to $n$-balls, EmEL++~\cite{mondal2021emel++}
extends this to richer constructors, ELBE~\cite{peng2022description}
and Box2EL~\cite{jackermeier2023box} use boxes so that conjunction and
existential restriction are handled geometrically, cone
embeddings~\cite{ozcep2023embedding} target $\mathcal{ALC}$, and
BoxE~\cite{abboud2020boxe} captures some inference patterns without
ontology-level subsumption. These models can in principle satisfy
levels above L0, but their training losses are still binary, positive
versus corrupted, and disjointness is not separately regularised. Our
framework predicts, and the experiments below test, that the additional
logical structure helps with MLF more than with PLF: ranking
\emph{within} $\Sund$ remains essentially unconstrained.

The term \emph{faithfulness} already appears in this literature, but in
a different, representation-level sense: an embedding is faithful if, at
zero training loss, it \emph{is} a model of the ontology and reproduces
its entailments. This is the sense of BoxEL~\cite{xiong2022faithful},
the strong-faithfulness results for
$\mathcal{ELH}$~\cite{lacerda2023strong}, and the semantic analysis of
\cite{bourgaux2024embeddings}; a query-level variant asks an embedding
to reproduce deductive query answers~\cite{sun2020faithful}, while in
interpretability ``faithful'' instead means an explanation reflects a
model's computation~\cite{jacovi2020towards}. Our criterion is
orthogonal and ranking-level: it constrains the \emph{order} a scoring
function induces over candidate axioms, not whether the embedding
realises a model, and a representation-faithful embedding may still
rank an undetermined axiom no higher than a contradictory one.

Probabilistic description logics~\cite{lukasiewicz2008managing}, the
distribution semantics for logic
programs~\cite{sato1995statistical}, and inconsistency-tolerant
query answering~\cite{lembo2010inconsistency} attach probabilities or
possibilistic weights to axioms. Our PLF criterion uses a related
object, the relative model count, in a different role: not as a
semantics the KB itself carries, but as a \emph{ground-truth target}
for evaluating embedding-based scoring functions on a classical KB.

Model counting (\#SAT) quantifies solution
spaces~\cite{gomes2021model} and, in weighted form, underlies
probabilistic inference~\cite{chavira2008probabilistic}. The relative
model count of Section~\ref{sec:modelcount} is the random-worlds degree
of belief of \cite{bacchus1996statistical}; our contribution is to use
it as the \emph{evaluation target} for embedding-based KBC, which to
our knowledge is new.

\begin{sloppypar}
Neuro-symbolic frameworks such as
DeepProbLog~\cite{manhaeve2018deepproblog},
NeurASP~\cite{yang2023neurasp}, Logic Tensor
Networks~\cite{badreddine2022logic}, and Logical Neural
Networks~\cite{riegel2020logical} couple differentiable models with
declarative logic; because our hierarchy is defined on the scoring
function $f$ alone, any such model can be placed on it
(Appendix~\ref{app:nesy}).
\end{sloppypar}

\section{Preliminaries}
\label{sec:prelim}

\subsection{Description logics}
\label{sec:dl}

Description logics (DLs) are logic-based knowledge representation
formalisms that provide the semantics of the Web Ontology Language
(OWL)~\cite{horrocks2005owl}; most are decidable fragments of
first-order logic, often within the two-variable fragment with counting
$\mathcal{C}^2$~\cite{van2008handbook}.

A DL ontology is defined over a signature $\Sigma = (N_C, N_R, N_I)$
consisting of disjoint sets of concept names $N_C$ (unary predicates),
role names $N_R$ (binary predicates), and individual names $N_I$
(constants)~\cite{dl_handbook}. Complex concepts are constructed
recursively from these primitives. The logic
$\mathcal{EL}$~\cite{baader2005pushing}, used in our experiments, is
a tractable fragment that allows top ($\top$), conjunction
($\sqcap$), and existential restriction ($\exists r.C$); subsumption
checking in $\mathcal{EL}$ runs in polynomial
time~\cite{baader2005pushing}.

An interpretation $\mathcal{I} = (\Delta^\mathcal{I},
\cdot^\mathcal{I})$ maps individuals to elements of a non-empty domain,
concept names to subsets, and role names to binary relations, extending
to complex concepts inductively in the standard way~\cite{dl_handbook}.
A knowledge base $\mathcal{O} = (\mathcal{T}, \mathcal{R},
\mathcal{A})$ comprises a non-empty TBox $\mathcal{T}$ of terminological axioms
(GCIs $C \sqsubseteq D$, equivalences $C \equiv D$), an RBox
$\mathcal{R}$ governing role characteristics, and an ABox
$\mathcal{A}$ of assertions $C(a)$ and $r(a,b)$. An interpretation is
a \emph{model} of $\mathcal{O}$ if it satisfies every axiom; we write
$\Mod(\mathcal{O})$ for the set of models of $\mathcal{O}$, and we
assume throughout that $\mathcal{O}$ is consistent
(i.e.\ $\Mod(\mathcal{O}) \neq \emptyset$).

\subsection{Knowledge base completion}
\label{sec:kbc}

Knowledge base completion (KBC) extends KGC link prediction to the
expressive power of description logics. While KGC predicts missing
edges (triples $(h, r, t)$, equivalent to ABox role
assertions)~\cite{liang2024survey,peng2023knowledge}, KBC predicts
missing logical axioms that may include complex concepts and
schema-level constraints~\cite{baader2006completing,lambrix2015completing,kulmanov2019embeddings}.

Formally, a \emph{candidate axiom} $\alpha$ is any well-formed formula
over $\Sigma$, and a KBC method learns a scoring function
$f:\mathcal{L}\times\mathcal{K}\to\mathbb{R}$ (axioms $\times$
knowledge bases) used to rank candidate extensions. Unlike standard
KGC, $\alpha$ may be an ABox assertion ($C(a)$, $r(a,b)$) or a TBox
axiom ($C \sqsubseteq D$, $C \equiv D$, or a disjointness
$C \sqcap D \sqsubseteq \bot$).

\section{The logical status of candidate axioms and relative model counting}
\label{sec:foundations}

This section establishes the two ingredients the faithfulness criteria
are built on: the three-way logical status of a candidate axiom under
the open world assumption, and a graded, computable refinement of that
status given by relative model counting.

\subsection{The logical status of candidate axioms}
\label{sec:trichotomy}

Because KBC operates under the open world assumption with full
deductive closure, the entailment status of a candidate axiom is not
binary. We adopt the following standard trichotomy~\cite{dl_handbook}.

\begin{definition}[Logical status under OWA]
\label{def:trichotomy}
For a consistent $\mathcal{O}$, every candidate axiom $\alpha$ is
exactly one of:
\emph{entailed} (provably true), $\mathcal{O}\models\alpha$, iff
$\Mod(\mathcal{O})\subseteq\Mod(\alpha)$;
\emph{contradictory} (provably false), $\mathcal{O}\models\neg\alpha$
(equivalently $\mathcal{O}\cup\{\alpha\}\models\bot$), iff
$\Mod(\mathcal{O})\cap\Mod(\alpha)=\emptyset$;
or \emph{undetermined} (contingent), iff $\mathcal{O}\not\models\alpha$
and $\mathcal{O}\not\models\neg\alpha$, equivalently both
$\Mod(\mathcal{O})\cap\Mod(\alpha)\neq\emptyset$ and
$\Mod(\mathcal{O})\setminus\Mod(\alpha)\neq\emptyset$.
\end{definition}

The set of candidate axioms then partitions into three disjoint
classes
$\Sent(\mathcal{O})$, $\Scon(\mathcal{O})$, and $\Sund(\mathcal{O})$;
when $\mathcal{O}$ is fixed by context we drop the argument. Standard
KGC training generates negatives by uniform corruption from
$\Scon\cup\Sund$ and suppresses them without distinction; fixing this
conflation is the premise of the paper.

\subsection{Bounded-domain model counting}
\label{sec:modelcount}

Let $\mathcal{O}$ be over signature
$\Sigma=(N_C,N_R,N_I)$ with $|N_I|=n$, and fix a finite domain
$\Delta=\{a_1,\ldots,a_n\}$ identified with the named individuals.
The \emph{Herbrand base} of $\Sigma$ over $\Delta$ is
\[
\mathcal{HB}(\Sigma,\Delta) = \bigcup_{C\in N_C}\{C(a_i)\}_{i=1}^n
\;\cup\; \bigcup_{r\in N_R}\{r(a_i,a_j)\}_{i,j=1}^n,
\]
and a \emph{Herbrand interpretation} is any subset
$I\subseteq\mathcal{HB}(\Sigma,\Delta)$. We write
$\Mod_H(\mathcal{O})$ for the set of Herbrand interpretations
satisfying every axiom of $\mathcal{O}$, with concept-language
constructors interpreted in the standard way over the finite domain
$\Delta$~\cite{bacchus1996statistical} and metalevel negation as set
complement in $\mathcal{HB}$: $\neg A(a) \equiv a\notin A^I$.

Restricting to Herbrand interpretations on the named individuals is a
domain-closure assumption that differs from the open-domain Tarskian
semantics of DLs; we adopt it because it makes $\#(\mathcal{O})$ finite
and well-defined while preserving satisfiability for fragments with the
finite model property ($\mathcal{EL},\mathcal{ALC}$, and others) at
sufficiently large $|\Delta|$, and because the resulting relative count
is a plausibility \emph{target}, not a redefinition of entailment,
which we still take classically.

\subsection{Relative model count}
\label{sec:relcount}

Define $\#(\mathcal{O}) = |\Mod_H(\mathcal{O})|$ and, for any
candidate axiom $\alpha$,
\begin{equation}
\label{eq:relcount}
P(\alpha\mid\mathcal{O}) \;=\;
\frac{\#(\mathcal{O}\cup\{\alpha\})}{\#(\mathcal{O})} \;\in\; [0,1].
\end{equation}

\begin{proposition}[Boundary conditions]
\label{prop:boundary}
For any consistent $\mathcal{O}$ and any candidate $\alpha$, we have
$P(\alpha\mid\mathcal{O}) = 1$ iff
$\Mod_H(\mathcal{O})\subseteq\Mod_H(\alpha)$ (Herbrand entailment);
$P(\alpha\mid\mathcal{O}) = 0$ iff
$\Mod_H(\mathcal{O})\cap\Mod_H(\alpha)=\emptyset$ (Herbrand
contradiction); and $P(\alpha\mid\mathcal{O}) \in (0,1)$ iff $\alpha$ is
Herbrand-undetermined w.r.t.\ $\mathcal{O}$.
\end{proposition}

The proof is immediate from \eqref{eq:relcount}. For DL fragments
with the finite model property and $|\Delta|$ taken large enough
(polynomial in $|\mathcal{O}|$ for $\mathcal{EL}$~\cite{baader2005pushing}),
Herbrand entailment coincides with classical entailment, so the
boundary conditions also hold for the trichotomy
$(\Sent,\Scon,\Sund)$ as defined in Section~\ref{sec:trichotomy}.

Inside $\Sund$, $P(\alpha\mid\mathcal{O})\in(0,1)$ grades how
constraining $\alpha$ is: near $1$ it holds in almost every model
(``almost entailed''), near $0$ in only a sliver (``almost
contradictory''). PLF requires $f$ to respect this ranking.

Exact model counting is \#P-complete~\cite{gomes2021model}, so the
tractability lever is the size of the bounded Herbrand domain rather
than the counting algorithm. Appendix~\ref{app:counting} details the
exact and approximate counting regimes and works a complete example:
$\mathcal{O}=\{A\sqsubseteq B\}$ over $|\Delta|=2$ gives
$\#(\mathcal{O})=576$, and adding $\exists r.A\sqsubseteq C$ gives
$\#(\mathcal{O}^{+})=425$, hence $P=425/576\approx 0.74$.

\section{A hierarchy of faithfulness criteria}
\label{sec:hierarchy}

Throughout this section, fix a consistent KB $\mathcal{O}$ and a
scoring function $f(\cdot,\mathcal{O}):\mathcal{L}\to\mathbb{R}$
provided by some completion model. Higher scores indicate higher
predicted plausibility. We define four progressively stronger
properties of $f$.

\subsection{The four criteria}

We define four criteria of increasing strictness, each a property of the
scoring function $f$ evaluated on the trichotomy
$(\Sent,\Scon,\Sund)$. Each criterion strictly implies the one before
it, as Theorem~\ref{thm:hierarchy} makes formal.

\begin{definition}[Discrimination]
$f$ is \emph{discriminative} on $\mathcal{O}$ iff for all
$\alpha^{+}\in\Sent$ and $\alpha^{-}\in\Scon$,
$f(\alpha^{+},\mathcal{O}) > f(\alpha^{-},\mathcal{O})$.
\end{definition}

Classical KGC metrics (MRR, Hits@$k$) approximate this and are silent
about $\Sund$.

\begin{definition}[Logical admissibility]
\label{def:adm}
$f$ is \emph{logically admissible} on $\mathcal{O}$ iff no contradiction
outranks any consistent axiom, i.e.\ for all $\alpha^{-}\in\Scon$ and
all $\beta\in\Sent\cup\Sund$,
$f(\alpha^{-},\mathcal{O}) < f(\beta,\mathcal{O})$; equivalently,
$\max_{\alpha^{-}\in\Scon} f(\alpha^{-},\mathcal{O}) <
\min_{\beta\in\Sent\cup\Sund} f(\beta,\mathcal{O})$.
\end{definition}

Equivalently, there is a threshold $\theta$ at which the accepted set
$\{\alpha : f(\alpha,\mathcal{O}) \geq \theta\}$ contains every
consistent candidate and no contradiction: a practitioner taking the
top-ranked axioms for review gets \emph{no contradictions} among them.
Admissibility is strictly stronger than discrimination, which
constrains only $\Sent$ versus $\Scon$ and is silent on $\Sund$. We
report the empirical $\mathrm{Adm@}k$ in
Section~\ref{sec:experiments}.

\begin{definition}[Monotonic logical faithfulness, MLF]
\label{def:mlf}
\sloppy
$f$ is \emph{monotonically logically faithful} on $\mathcal{O}$ iff
for every $(\alpha^{-},\alpha^{?},\alpha^{+})$ in
$\Scon\times\Sund\times\Sent$,
\[
f(\alpha^{-},\mathcal{O}) < f(\alpha^{?},\mathcal{O}) <
f(\alpha^{+},\mathcal{O}).
\]
\end{definition}

MLF is strictly stronger than admissibility: admissibility only keeps
$\Scon$ below the consistent axioms as a block, whereas MLF inserts
$\Sund$ as a distinct middle band ordered strictly below $\Sent$.

MLF is purely ordinal across classes and imposes no order \emph{within}
$\Sund$, yet two undetermined axioms can differ sharply in
plausibility: one holding in $999{,}000$ of a KB's million models is
``almost entailed'', one holding in only $1{,}000$ ``almost
contradictory''. The relative model count captures this.

\begin{definition}[Probabilistic logical faithfulness, PLF]
\label{def:plf}
$f$ is \emph{probabilistically logically faithful} on $\mathcal{O}$
iff for every pair of candidate axioms $\alpha_1, \alpha_2$,
\[
P(\alpha_1\mid\mathcal{O}) > P(\alpha_2\mid\mathcal{O}) \;\Longrightarrow\;
f(\alpha_1,\mathcal{O}) > f(\alpha_2,\mathcal{O}),
\]
where $P(\alpha\mid\mathcal{O}) =
\#(\mathcal{O}\cup\{\alpha\})/\#(\mathcal{O})$ is the relative model
count over a fixed finite-domain semantics
(Section~\ref{sec:modelcount}).
\end{definition}

PLF is strictly stronger than MLF: it is silent only on pairs with
identical relative model counts (and there is no $f$-constraint
within $\Sent$ or within $\Scon$ themselves, where the count is
constant at 1 and 0 respectively).

\subsection{The hierarchy theorem}

The four criteria differ in operational feasibility. L0--L2 are
testable on any ontology whose entailment is decidable, using
reasoner-generated $\Sent$, $\Sund$, $\Scon$ samples; L3 additionally
requires the $\#P$-complete relative model count
$P(\alpha\mid\mathcal{O})$, so we treat it as a benchmark criterion
rather than a runtime property (Section~\ref{sec:experiments}).

\begin{theorem}[Strict hierarchy]
\label{thm:hierarchy}
Let $\mathcal{O}$ be consistent, assume the relative model count
$P(\cdot\mid\mathcal{O})$ is well-defined (Section~\ref{sec:modelcount}),
and assume $\Sent$, $\Scon$, and $\Sund$ are all non-empty (otherwise the
criteria degenerate).
Then
\[
\mathrm{PLF} \;\Longrightarrow\; \mathrm{MLF} \;\Longrightarrow\;
\mathrm{Logical\;Admissibility} \;\Longrightarrow\;
\mathrm{Discrimination},
\]
and none of the implications can be reversed: there exist scoring
functions satisfying any one level but failing the next stronger one.
\end{theorem}

A proof is given in Appendix~\ref{app:proofs}:
$\mathrm{PLF}\Rightarrow\mathrm{MLF}$ follows from the boundary values
$P(\alpha^{-})=0$, $P(\alpha^{+})=1$ and $0 < P(\alpha^{?}) < 1$; the
remaining forward implications are ordinal, and the non-reversals use
small explicit constructions. See Appendix~\ref{app:example} for a working example.

\section{Experimental evaluation}
\label{sec:experiments}

\subsection{Setup}

We organise the evaluation around three questions. RQ1: do existing
KGC and KBC models satisfy MLF, and at what rate do they violate the
ordering across $\Sent,\Sund,\Scon$? RQ2: conditional on L0, how large
is the residual confusion of $\Sund$ with $\Scon$, the failure mode our
critique targets? RQ3: on fragments where $P(\alpha\mid\mathcal{O})$ is
exactly computable, do logic-geometric models satisfy PLF, i.e.\ rank
$\Sund$ by relative model count?

We use two ontologies spanning the model-counting tractability
spectrum: the Pizza fragment, an $\mathcal{EL}$ version of the
well-known Pizza ontology small enough that $\#(\mathcal{O})$ is
feasible to compute exactly via \#SAT (RQ3); and the GO-plus slice, a
connected slice of the Gene Ontology (go-plus) of $O(10^3)$ classes, whose
trichotomy is computed via ELK~\cite{elk} and whose relative model
counts are computed exactly with \textsc{pyganak} (\textsc{Ganak}) over
the same bounded Herbrand domain (RQ1--RQ3). We use bounded Herbrand
domain size $|\Delta|=2$ for both ontologies, matching
Appendix~\ref{app:counting}'s worked example: the smallest domain
admitting non-trivial role instantiation while keeping exact \#SAT
tractable over the full test suite. The boundary conditions of
Proposition~\ref{prop:boundary} hold at this size for the $\mathcal{EL}$
axioms considered, and PLF requires only the ranking of $P$-values on
$\Sund$, which is stable across $|\Delta|$ choices.

We evaluate geometric KGC models
(TransE~\cite{bordes2013translating}, DistMult, ComplEx), a hybrid
model (BoxE~\cite{abboud2020boxe}), and logic-geometric KBC models
(ELEmbeddings~\cite{kulmanov2019embeddings},
EmEL++~\cite{mondal2021emel++}, Box2EL~\cite{jackermeier2023box}).

We use the ELEmbeddings and EmEL++ losses as in their reference
implementations, which deviate in two normal forms from the formulas as
originally published.\footnote{The released code omits the
superclass-radius term of the conjunctive-subsumption loss (Eq.~(2) of
\cite{kulmanov2019embeddings}) and uses a corrected loss for
$\exists r.C\sqsubseteq D$; Appendix~\ref{app:elem} gives the details.
We report the implementations as released and widely used.} We therefore do not
attribute the conjunctive-subsumption behaviour of ball-based models to
geometry alone, and anchor the cross-model comparison on the unaffected
box-based and translational models.

We sample test sets $T_{+}\subset\Sent$, $T_{?}\subset\Sund$,
$T_{-}\subset\Scon$ via the reasoner and report one metric per level. For the GO-plus slice, no negatives were generated for GCI1, GCI2, and GCI3 normal forms, so we report full metrics only for GCI0 and GCI1$_\bot$ settings in this case. 
$\mathrm{AUC}_{+/-}$, the AUC of $f$ separating $T_{+}$ from $T_{-}$,
tests L0; $\mathrm{Adm}@k$, the fraction of the top $k$ not in $\Scon$,
averaged over ranking queries, tests L1; the faithfulness violation rate $\mathrm{FVR}=
\Pr_{T_{-}\times T_{?}\times T_{+}}[\neg(f(\alpha^{-})<f(\alpha^{?})<
f(\alpha^{+}))]$ tests L2, with the directional
$\text{Pair-FVR}=\Pr_{T_{-}\times T_{?}}[f(\alpha^{?})\leq
f(\alpha^{-})]$ isolating the $\Sund$-versus-$\Scon$ confusion; and the
Spearman correlation $\rho_P$ between $f$ and $P(\cdot\mid\mathcal{O})$
on $T_{?}$ tests L3. Standard MRR and Hits@$k$ on $T_{+}$ are a
ranking-accuracy sanity check. In all reported tables, column headers
are marked $\uparrow$ (higher is better) or $\downarrow$ (lower is
better).

For each model and dataset pair we train on $\mathcal{A}$ (and
$\mathcal{T}$ where the model accepts it), evaluate on the three
classes above, and report all metrics. Ontology sizes and the
per-normal-form test-set composition are given in
Appendix~\ref{app:datastats}. Each configuration is trained
over multiple random seeds; the body tables report the mean and the standard deviation across
seeds, full results are contained in Appendix~\ref{app:results}. The effects we report are large
relative to this seed-to-seed variation, which is small throughout
(standard deviations are mostly below $0.05$). The training-time
intervention sweep of Section~\ref{sec:training} is the one experiment
we still run at a single seed, as noted there.

\subsection{Results}
\label{sec:results}

\begin{table}[h!]
\centering
\begin{tabular}{lrrrrr}
\toprule
Model & $\mathrm{AUC}_{+/-}$ $\uparrow$ & Pair-FVR $\downarrow$ & $\rho_P$ $\uparrow$ & MRR $\uparrow$ & Hits@10 $\uparrow$ \\
\midrule
\multicolumn{6}{l}{\emph{Logic-geometric (KBC)}}\\
ELEmbeddings & 0.70 $\pm$ 0.01 & 0.48 $\pm$ 0.00 & \phantom{-}0.62 $\pm$ 0.02 & 0.14 $\pm$ 0.00 & 0.42 $\pm$ 0.01 \\
EmEL++       & 0.73 $\pm$ 0.01 & 0.49 $\pm$ 0.00 & \phantom{-}0.63 $\pm$ 0.05 & 0.15 $\pm$ 0.01 & 0.41 $\pm$ 0.04 \\
Box2EL       & 0.74 $\pm$ 0.01 & 0.60 $\pm$ 0.01 & \phantom{-}0.53 $\pm$ 0.00 & 0.17 $\pm$ 0.01 & 0.51 $\pm$ 0.01 \\
\midrule
\multicolumn{6}{l}{\emph{Geometric (KGC)}}\\
BoxE     & 0.42 $\pm$ 0.06 & 0.73 $\pm$ 0.02 & \phantom{-}0.16 $\pm$ 0.04 & 0.10 $\pm$ 0.00 & 0.23 $\pm$ 0.04 \\
ComplEx  & 0.53 $\pm$ 0.03 & 0.45 $\pm$ 0.01 & -0.13 $\pm$ 0.03 & 0.08 $\pm$ 0.01 & 0.11 $\pm$ 0.01 \\
DistMult & 0.52 $\pm$ 0.09 & 0.47 $\pm$ 0.08 & -0.03 $\pm$ 0.08 & 0.05 $\pm$ 0.02 & 0.08 $\pm$ 0.06 \\
TransE   & 0.61 $\pm$ 0.04 & 0.61 $\pm$ 0.05 & \phantom{-}0.18 $\pm$ 0.07 & 0.20 $\pm$ 0.00 & 0.32 $\pm$ 0.03 \\
\bottomrule
\end{tabular}
\caption{Faithfulness across models on atomic subsumption
($C\sqsubseteq D$, GCI0) over the Pizza ontology. Higher
$\mathrm{AUC}_{+/-}$ and $\rho_P$ are better; Pair-FVR near $0.5$ means
$\Sund$ is not separated from $\Scon$. Ranking accuracy (MRR, Hits@10)
does not predict faithfulness. Full metrics, the GO-plus slice, and all
normal forms are in Appendix~\ref{app:results}.}
\label{tab:models}
\end{table}

The one normal form on which the ball-based models invert rather than
merely underperform is conjunctive subsumption (GCI1), where
ELEmbeddings and EmEL++ fall well below chance ($\mathrm{AUC}_{+/-}$
near $0.02$--$0.05$); an AUC this close to $0$ signals near-total
inversion, with the loss ordering entailments and contradictions almost
perfectly backwards rather than merely failing to separate them. These
are exactly the two models whose released loss for
$C\sqcap D\sqsubseteq E$ ignores the size of the right-hand concept
(Appendix~\ref{app:elem}), so a contradictory axiom with a centrally
placed superclass can outscore an entailed one. Box2EL, which represents
conjunction through the exact intersection of axis-aligned boxes, is far
less affected ($\mathrm{AUC}_{+/-}=0.39$), so the collapse tracks the
loss as implemented rather than an intrinsic limit of region-based
embeddings.

Two findings stand out. First, faithfulness is a distinct axis from
ranking accuracy: in Table~\ref{tab:models} the ordering of models by
MRR (TransE, Box2EL, ELEmbeddings) does not match their ordering by
Pair-FVR or $\rho_P$, and TransE attains the highest MRR yet scores
$\Sund$ no higher than $\Scon$. Logic-geometric models reach higher
$\mathrm{AUC}_{+/-}$ and $\rho_P$ than the geometric baselines, but on
atomic subsumption even the best leave Pair-FVR near $0.5$, so they do
not cleanly satisfy L2. Second, where a model sits on
the hierarchy depends on the logical construct
(Table~\ref{tab:constructs}, Appendix~\ref{app:results}): disjointness
is handled faithfully, conjunctive subsumption inverts, and atomic and
existential subsumption are only weakly separated. No setting reaches
$\rho_P$ close to $1$, so probabilistic logical faithfulness (L3) is met
by none of the current models, consistent with the absence of any
model-count signal in their training objectives.

\subsection{Training-time interventions}
\label{sec:training}

The hierarchy also indicates where to intervene during training. We
sketch three directions, specialising two existing lines,
type-constrained negative sampling~\cite{krompass2015type} and
rule-injected embeddings~\cite{guo2016jointly,guo2018knowledge}, to
the trichotomy.
\emph{Type-aware negative sampling} (L1--L2) replaces uniform
corruption with a stratified sampler that draws a fraction $p_\bot$ of
negatives from $\Scon$ via cheap reasoning shortcuts (asserted
disjointness, domain/range violations), introducing an explicit signal
that separates $\Scon$ from $\Sund$. Two further directions modify the
loss rather than the sampler: a \emph{contradiction margin loss} (L2)
that penalises scores of sampled contradictions below a threshold, and
an \emph{entailed-positives objective} (L0--L1) that adds entailed
axioms as further positives. None targets PLF directly, since ranking
within $\Sund$ would require the loss to be informed by relative model
counts, intractable for non-toy $\mathcal{O}$; the two loss-based
interventions are left for future work.

We evaluate the stratified sampler by sweeping the mixture for
each logic-geometric model on the Pizza ontology from $p_\bot=1$ (all
negatives from $\Scon$) to $p_\bot=0$ (all random corruptions), per
normal form (GCI0, GCI2, GCI1$_\bot$, GCI3; GCI1 excluded,
its failure driven by the released losses, Appendix~\ref{app:elem}). The
endpoint effect is uniform: in all 12 model--construct settings
(Figure~\ref{fig:sweep-all}),
training only on contradiction negatives yields lower FVR than training
only on random corruptions, from noise-level on GCI0 for the ball-based
models ($\leq 0.01$) to $0.13$ absolute for Box2EL on GCI3. A five-seed replication of the full sweep
(Appendix~\ref{app:results},
Tables~\ref{app:tab-sweep-b2el-pizza-gci0}--\ref{app:tab-sweep-emel-pizza-gci3})
confirms this endpoint contrast on every panel. The response is not
monotone in between, so the sweep supports the endpoint contrast,
not a calibrated mixing schedule. MRR mostly moves the
opposite way (6 of 12 settings), reproducing the
rank--faithfulness dissociation of Table~\ref{tab:models} as an
intervention within a single model. Even at $p_\bot=1$, FVR stays above
$0.43$ for most settings: the sampler supplies the missing
$\Scon$-versus-$\Sund$ signal but does not by itself make a model
faithful.

\section{Discussion and conclusion}
\label{sec:discussion}

The move from KGC to KBC is a change of evaluation target, not just of
dataset: once OWA and deductive closure are in scope, ranking observed
triples above random corruptions is no longer sufficient, and the
hierarchy says what \emph{is} required and at which strictness.

The four criteria are operationalized on two tiers. L0--L2 depend only
on the reasoner-computable trichotomy $(\Sent,\Sund,\Scon)$ and are
testable on any ontology whose entailment is decidable; they are
practitioner tools, applicable at deployment scale. L3 additionally
requires the relative model count $P(\alpha\mid\mathcal{O})$, which is
$\#P$-complete: exact via $\#$SAT on small ontologies, approximate via
ApproxMC (Appendix~\ref{app:counting}) on medium ones, and infeasible at
scale. This is a design property rather than a limitation: the hierarchy
orders the \emph{strength} of ordinal constraints on $f$, while
operational feasibility is a separate axis on which L3 sits at the
benchmark end, providing a reference criterion on ontologies where
$P(\alpha\mid\mathcal{O})$ is computable, against which models trained
on any ontology can be certified or falsified. The tension between the
bounded Herbrand semantics of $P$ and the open-domain semantics of DLs
is by construction: $P$ is a plausibility target, not a redefinition of
entailment (Section~\ref{sec:modelcount}), aligned with the classical
trichotomy on $\Sent$ and $\Scon$ by
Proposition~\ref{prop:boundary}. Our scope is $\mathcal{EL}$, with
$\mathcal{ALC}$/$\mathcal{SROIQ}$ and broader
$\mathcal{EL}/\mathcal{EL}^{++}$ validation as future work; sampling
$T_{-}$ requires disjointness axioms, absent from natural candidates
such as GALEN.

The hierarchy is defined on the scoring function alone, so any KBC
method can be placed on it, including neuro-symbolic frameworks whose
logical layer differs from the geometric embeddings evaluated here;
Appendix~\ref{app:nesy} places representative frameworks (DeepProbLog,
NeurASP, LTN, LNN) on the hierarchy and identifies what each would need
to reach MLF and PLF.

We have argued that knowledge base completion requires criteria that
respect the OWA trichotomy, and organised them into a hierarchy
grounded at the top in model counting. Empirically, current models
cluster near the bottom, satisfying discrimination but violating MLF;
where a model sits depends on the logical construct; and the
random-corruption sampler is itself implicated: contradiction negatives
improve faithfulness at a cost in ranking accuracy. The broader
programme is to make KBC \emph{about} logical correctness, not ranking
in a TBox-stripped graph.

\bibliographystyle{splncs04}
\bibliography{bibliography}

\appendix

\section{Model counting example (extended)}
\label{app:counting}

In our experiments we compute exact \#SAT
with \textsc{Ganak} (via \textsc{pyganak}) over a bounded domain for
both the Pizza fragment and the GO-plus slice, so an exact relative count is
available for every test axiom; the GO-plus slice is kept tractable by the
domain bound, not by approximation. For ontologies beyond this scale
two fallbacks are available but not used here: approximate counting
(\textsc{ApproxMC}, with multiplicative $(1\pm\varepsilon)$ error at
confidence $1-\delta$), and, when even that is infeasible, only the
trichotomy $(\Sent,\Scon,\Sund)$ via a DL reasoner such as ELK or
HermiT~\cite{glimm2014hermit,elk}, which suffices for MLF but not for
PLF beyond bounded samples.

We now work the example cited in Section~\ref{sec:relcount}.
Let $\mathcal{O}=\{A\sqsubseteq B\}$ over signature
$\Sigma=(N_C,N_R,N_I)$ with $N_C=\{A,B,C\}$, $N_R=\{r\}$,
$N_I=\{a_1,a_2\}$, and Herbrand domain
$\Delta=\{a_1,a_2\}$. The Herbrand base contains
\[
\mathcal{HB}(\Sigma,\Delta) = \{A(a_i),B(a_i),C(a_i)\}_{i=1,2}\;\cup\;
\{r(a_i,a_j)\}_{i,j=1,2},
\]
i.e.\ ten ground atoms.

First, we compute $\#(\mathcal{O})$. Reading $A\sqsubseteq B$ clausally
as $\forall i.\,\neg A(a_i)\lor B(a_i)$
yields two clauses, each over the two atoms
$\{A(a_i),B(a_i)\}$. Each clause has $3$ satisfying assignments out
of $4$, and the two clauses share no atoms, so there are $9$ valid
assignments to $\{A(a_1),A(a_2),B(a_1),B(a_2)\}$. The remaining six
atoms
\[
\{C(a_1),C(a_2),r(a_1,a_1),r(a_1,a_2),r(a_2,a_1),r(a_2,a_2)\}
\]
are unconstrained, contributing a factor $2^6=64$. Hence
$\#(\mathcal{O}) = 9\cdot 64 = 576$.

Next, we compute $\#(\mathcal{O}\cup\{\exists r.A\sqsubseteq C\})$.
The added axiom reads, again clausally,
\[
\forall i,j.\;\neg r(a_i,a_j)\lor \neg A(a_j)\lor C(a_i),
\]
i.e.\ four ground clauses, one per $(i,j)$. A direct enumeration over
the $9$ valid $(A,B)$-assignments and, for each, a count of the $r$-
and $C$-assignments that satisfy the four added clauses, gives a total
of $425$ models. The intermediate counts split by $A$-extension as
follows:

\begin{tabular}{lcc}
\toprule
$A$-assignment & $\#$ valid $(r,C)$ tuples & matching $B$-assignments
\\\midrule
$A^I=\emptyset$ & $2^4 \cdot 2^2 = 64$ & $1\cdot 4 = 4$ \\
$A^I=\{a_1\}$  & $\sum_{r}\prod_i \mathbf{1}[r(a_i,a_1)\Rightarrow C(a_i)]$
\quad$= 36$ & $1\cdot 2 = 2$ \\
$A^I=\{a_2\}$  & $36$ (symmetric) & $2$ \\
$A^I=\{a_1,a_2\}$ & $\sum_{r}\prod_i \mathbf{1}[\exists j: r(a_i,a_j) \Rightarrow C(a_i)]
= 25$ & $1$ \\
\bottomrule
\end{tabular}

\noindent
Multiplying matching $B$-assignments by $(r,C)$-tuples:
$4\cdot 64 + 2\cdot 36 + 2\cdot 36 + 1\cdot 25 = 256 + 72 + 72 + 25 = 425$.

The relative model count is therefore
\[
P(\exists r.A\sqsubseteq C \mid \mathcal{O}) = \frac{425}{576}
\approx 0.7378.
\]
Since $0 < 425/576 < 1$, by Proposition~\ref{prop:boundary} the
candidate axiom is Herbrand-undetermined relative to $\mathcal{O}$,
and a PLF-faithful model would rank it above any axiom $\beta\in\Sund$
with $P(\beta\mid\mathcal{O}) < 0.7378$ and below any $\beta'$ with
$P(\beta'\mid\mathcal{O}) > 0.7378$.

\section{Proof of Theorem~\ref{thm:hierarchy}}
\label{app:proofs}

Throughout, $\Sent$, $\Scon$, and $\Sund$ are non-empty as assumed in
Theorem~\ref{thm:hierarchy}. We prove the three forward implications in
turn, then exhibit counterexamples for the converses.

For PLF $\Rightarrow$ MLF, let $f$ be PLF and let
$(\alpha^{-},\alpha^{?},\alpha^{+})\in
\Scon\times\Sund\times\Sent$. By
Proposition~\ref{prop:boundary}, $P(\alpha^{-}\mid\mathcal{O})=0$,
$P(\alpha^{?}\mid\mathcal{O})\in(0,1)$,
$P(\alpha^{+}\mid\mathcal{O})=1$, so
$P(\alpha^{-}\mid\mathcal{O}) <
P(\alpha^{?}\mid\mathcal{O}) < P(\alpha^{+}\mid\mathcal{O})$.
Applying PLF to each of the two strict inequalities yields
$f(\alpha^{-}) < f(\alpha^{?}) < f(\alpha^{+})$, which is exactly MLF.

For MLF $\Rightarrow$ logical admissibility, let $f$ be MLF, fix any
$\alpha^{-}\in\Scon$ and any consistent $\beta\in\Sent\cup\Sund$, and
pick (using non-emptiness) some $\alpha^{?}\in\Sund$ and
$\alpha^{+}\in\Sent$. If $\beta\in\Sund$, the triple
$(\alpha^{-},\beta,\alpha^{+})$ gives $f(\alpha^{-})<f(\beta)$; if
$\beta\in\Sent$, the triple $(\alpha^{-},\alpha^{?},\beta)$ gives
$f(\alpha^{-})<f(\alpha^{?})<f(\beta)$. Either way
$f(\alpha^{-})<f(\beta)$ for all such $\alpha^{-},\beta$, i.e.\
$\max_{\Scon} f < \min_{\Sent\cup\Sund} f$, which is admissibility
(Definition~\ref{def:adm}).

For logical admissibility $\Rightarrow$ discrimination, admissibility
gives $f(\alpha^{-})<f(\beta)$ for every $\alpha^{-}\in\Scon$ and every
$\beta\in\Sent\cup\Sund$, in particular for every $\alpha^{+}\in\Sent$,
which is discrimination.

For non-reversal of PLF $\Rightarrow$ MLF, take $\mathcal{O}$ admitting
two undetermined axioms $\alpha_1,\alpha_2$
with $P(\alpha_1\mid\mathcal{O}) > P(\alpha_2\mid\mathcal{O})$. Define
$f$ to satisfy MLF and to set $f(\alpha_1)<f(\alpha_2)$ within
$\Sund$; this is consistent with MLF (which is silent within
$\Sund$) but violates PLF.

For non-reversal of MLF $\Rightarrow$ admissibility, take
$\alpha^{-}\in\Scon$, $\alpha^{?}\in\Sund$, $\alpha^{+}\in\Sent$ and
set $f(\alpha^{-})<f(\alpha^{+})<f(\alpha^{?})$. Then every
contradiction is below every consistent axiom (admissibility holds), but
MLF fails on $(\alpha^{-},\alpha^{?},\alpha^{+})$ because
$f(\alpha^{?})>f(\alpha^{+})$.

For non-reversal of admissibility $\Rightarrow$ discrimination, take
$\alpha^{-}\in\Scon$, $\alpha^{?}\in\Sund$, $\alpha^{+}\in\Sent$ and
set $f(\alpha^{?})<f(\alpha^{-})<f(\alpha^{+})$. Then
$f(\alpha^{+})>f(\alpha^{-})$ (discrimination holds), but the
contradiction $\alpha^{-}$ outranks the consistent axiom $\alpha^{?}$,
so admissibility fails.

\hfill$\square$

\section{Implementation deviations and corrections}
\label{app:elem}

The ELEmbeddings and EmEL++ baselines are run with their reference
implementations. For two normal forms these implementations differ from
the loss functions in the original paper~\cite{kulmanov2019embeddings},
and the deviation for the conjunctive normal form is the proximate cause
of the GCI1 inversion reported in Section~\ref{sec:results}. We record
these deviations, together with a correction to the normalisation
toolchain, here for transparency. Throughout, a concept $C$ is an
$n$-ball with centre $c$ and radius $r_C$, $\gamma$ is the margin, and
the normalisation and regularisation terms are elided.

\paragraph*{Conjunctive subsumption $C\sqcap D\sqsubseteq E$ (GCI1, the
NF2 normal form).}
Equation~(2) of \cite{kulmanov2019embeddings} defines four penalty
terms:
\begin{align*}
\ell_{C\sqcap D\sqsubseteq E}
 &= \max(0,\ \lVert c-d\rVert - r_C - r_D - \gamma) \tag{i}\\
 &\quad + \max(0,\ \lVert c-e\rVert - r_C - \gamma) \tag{ii}\\
 &\quad + \max(0,\ \lVert d-e\rVert - r_D - \gamma) \tag{iii}\\
 &\quad + \max(0,\ \min(r_C,r_D) - r_E - \gamma). \tag{iv}
\end{align*}
Term~(i) forces the balls of $C$ and $D$ to overlap; terms~(ii)
and~(iii) pull the centre of $E$ inside both balls, that is, into the
lens $C\sqcap D$; and term~(iv), the only term involving the radius
$r_E$ of the right-hand concept, forces $E$ to be at least as large as
the smaller of $C$ and $D$, which is what makes the intersection lie
\emph{within} $E$ rather than merely overlap it. The released
implementation drops term~(iv) and keeps only (i)--(iii); the radius
$r_E$ of the right-hand concept never enters the loss at all (the code
fetches the radii of $C$ and $D$ but never reads the radius of $E$).
Consequently the score of
$C\sqcap D\sqsubseteq E$ is independent of the size of $E$: it rewards
any configuration in which the centre of $E$ falls inside the lens,
regardless of whether the lens is actually contained in $E$. A
contradictory axiom whose right-hand concept happens to be large or
centrally placed then receives a lower (more plausible) loss than a
genuinely entailed axiom whose right-hand concept is specific and
off-centre. Because the evaluation score is the training loss itself
(the model has no separate scorer), this surrogate is what the
discrimination AUC measures, and it is anti-correlated with entailment
on the GCI1 test sets, driving $\mathrm{AUC}_{+/-}$ below $0.5$ for
ELEmbeddings and EmEL++. The defect is shared by the original authors'
released code, where term~(iv) appears commented out; it is therefore a
property of the implementation as used by the community, not of our
re-implementation. Box-based models such as Box2EL represent
$C\sqcap D$ as the exact intersection of axis-aligned boxes and do not
exhibit the near-total inversion of the ball-based models, though their
$\mathrm{AUC}_{+/-}$ on GCI1 still sits somewhat below $0.5$.

Because $\mathrm{AUC}_{+/-}$ is a rank statistic, a value near $0$ is
the sign-flip of a value near $1$ ($\mathrm{AUC}_{+/-}(-f)=1-
\mathrm{AUC}_{+/-}(f)$ exactly, ties included), so one might be tempted
to simply negate the GCI1 score and recover a near-perfect classifier.
This does not yield a faithful model. The negation would have to be
applied to GCI1 alone (the other normal forms already score above
$0.5$), which presupposes the symbolic normal form the score is meant
to be agnostic to, and the discriminating signal is the size and
placement of the right-hand concept rather than entailment itself, a
feature of how the contradiction set is sampled. The principled fix is
to restore term~(iv) so the loss penalises an undersized $E$ directly,
recovering separation in the correct direction rather than reading the
inverted surrogate backwards.

For comparison, the NF1 loss for atomic subsumption $C\sqsubseteq D$
(Equation~(1) of the same paper),
$\ell_{C\sqsubseteq D}=\max(0,\ \lVert c-d\rVert + r_C - r_D - \gamma)$,
does use the superclass radius $r_D$, so the omission is specific to the
conjunctive normal form.

\paragraph*{Existential subsumption $\exists r.C\sqsubseteq D$ (GCI3,
the NF4 normal form).}
Here the implemented loss \emph{corrects} the published formula. The
reference code uses
$\ell_{\exists r.C\sqsubseteq D}=\max(0,\ \lVert c-r-d\rVert + r_C - r_D
- \gamma)$, a containment-style margin $+r_C-r_D$ matching the NF3 loss
for $C\sqsubseteq\exists r.D$, whereas Equation~(4) of the original
paper prints $-r_C-r_D$, which encodes a disjointness-like rather than a
subsumption constraint. We use the corrected implementation; unlike the
GCI1 case this deviation makes the loss more, not less, faithful to the
intended semantics.

We report all EL embedding results with the implementations as released
and widely used. The two deviations are confined to the NF2 and NF4
normal forms; the box-based (Box2EL, BoxE) and translational (TransE,
DistMult, ComplEx) baselines are unaffected, and the cross-model
comparison in Section~\ref{sec:results} is anchored on them.

\paragraph*{Normalisation (\texttt{jcel}).}
The GCI normal-form splits (GCI0 to GCI3, and the disjointness form) are
produced by normalising each ontology into $\mathcal{EL}^{++}$ normal
form with \texttt{jcel}~\cite{mendez2012jcel}. We found and fixed a bug
in \texttt{jcel}'s normaliser that produced incorrect normalised axioms;
left uncorrected it would have corrupted the normal-form train and test
sets and, through them, the deductive closure and the relative model
counts that define the trichotomy. All results reported here use the
corrected normaliser.

\section{Working example}
\label{app:example}
\begin{example}
Let $\mathcal{O}$ have $\mathcal{T}=\{\text{PhDStudent}\sqsubseteq
\text{Student},\;\text{Student}\sqcap\text{Professor}\sqsubseteq\bot\}$
and $\mathcal{A}=\{\text{PhDStudent}(\text{alice})\}$. For ABox
candidates, $\text{Student}(\text{alice})\in\Sent$,
$\text{Professor}(\text{alice})\in\Scon$ (disjoint with the entailed
$\text{Student}(\text{alice})$), and
$\text{TeachingAssistant}(\text{alice})\in\Sund$, so a faithful $f$
must order $f(\text{Professor})<f(\text{TA})<f(\text{Student})$. The
same holds for TBox candidates:
$\text{PhDStudent}\sqsubseteq\text{Student}\sqcup\text{Professor}
\in\Sent$ (already implied),
$\text{PhDStudent}\sqsubseteq\text{Professor}\in\Scon$ (it would force
the disjoint $\text{Student}$ and $\text{Professor}$ to share
$\text{alice}$), and
$\text{TeachingAssistant}\sqsubseteq\text{Student}\in\Sund$.
\end{example}

\section{Placing neuro-symbolic frameworks on the hierarchy}
\label{app:nesy}

The hierarchy is defined on the scoring function $f$ alone
(Section~\ref{sec:hierarchy}), so any KBC method can be placed on it,
including neuro-symbolic frameworks whose logical layer differs
substantially from the geometric embeddings evaluated in
Section~\ref{sec:experiments}. Frameworks with an \emph{exact logical
layer} (DeepProbLog~\cite{manhaeve2018deepproblog},
NeurASP~\cite{yang2023neurasp}) can in principle reach MLF: their
inference layers enforce zero probability on contradictions and unit
probability on entailed axioms, aligning with the boundary conditions
of Proposition~\ref{prop:boundary}. Reaching PLF additionally requires
the inference to align with the relative model count; DeepProbLog's
weighted model counting is the closest existing mechanism and would in
principle target $\rho_P$ directly, though translating DL axioms with
existential restrictions into ProbLog programs is non-trivial.
Frameworks with a \emph{fuzzy or weighted logical layer}
(LTN~\cite{badreddine2022logic}, LNN~\cite{riegel2020logical}) reach
L0 by construction, since fuzzy satisfaction losses train $f$ to
distinguish satisfied from violated axioms. MLF depends on whether
contradictions produce truth collapse rather than intermediate fuzzy
scores, which the disjointness axioms can enforce in principle. PLF is
out of reach without an additional model-counting mechanism, since
fuzzy or interval-based semantics do not compose into a probability
measure over models. These are ceiling placements determined by the
logical layer's expressivity; empirical placement additionally depends
on the training regime and on data coverage of contradictory and
undetermined axioms. Full experimental placement of these frameworks on
$\rho_P$ is a direct extension of this work: our released evaluation
pipeline accepts any scoring function $f$, so the barrier is
engineering rather than framework design.

\section{Dataset statistics}
\label{app:datastats}

Table~\ref{app:tab-datastats} reports the size of each ontology and the
composition of the evaluation splits. Axiom counts are by normal form,
excluding trivial axioms (reflexive $C\sqsubseteq C$,
$C\sqsubseteq\top$, $\bot\sqsubseteq C$, and $C\sqsubseteq\bot$). For
each split, $T_{+}$, $T_{-}$, and $T_{?}$ are the reasoner-generated
entailed, contradictory, and undetermined candidate sets used to
evaluate L0--L3; the domain size $|\Delta|$ is the bounded Herbrand
domain used for model counting (Section~\ref{sec:experiments}). The GO
slice is evaluated with a full trichotomy on the two normal forms for
which a contradictory set is available (GCI0 and GCI1$_\bot$); for GCI1,
GCI2, and GCI3 no contradictory set is generated, so only $T_{+}$ and
$T_{?}$ are defined and the reduced metrics ($\rho_P$, MRR, Hits@10) are
reported. Training-axiom counts are reported for the GCI0 split.

\begin{table}[h]
\centering
\begin{tabular}{lrr}
\toprule
& Pizza & GO-plus \\
\midrule
Classes $|N_C|$ & 106 & 653 \\
Roles $|N_R|$ & 4 & 6 \\
Domain size $|\Delta|$ & 2 & 2 \\
\midrule
\multicolumn{3}{l}{\emph{Training axioms by normal form (GCI0 split)}}\\
GCI0 \, $C\sqsubseteq D$ & 62 & 546 \\
GCI1 \, $C\sqcap D\sqsubseteq E$ & 6 & 18 \\
GCI1$_\bot$ \, $C\sqcap D\sqsubseteq\bot$ & 398 & 29 \\
GCI2 \, $C\sqsubseteq\exists R.D$ & 153 & 384 \\
GCI3 \, $\exists R.C\sqsubseteq D$ & 9 & 18 \\
\midrule
\multicolumn{3}{l}{\emph{GCI0 evaluation split}}\\
$|T_{+}|$ (entailed) & 97 & 349 \\
$|T_{-}|$ (contradictory) & 914 & 38 \\
$|T_{?}|$ (undetermined) & 988 & 999 \\
\midrule
\multicolumn{3}{l}{\emph{GCI1$_\bot$ evaluation split}}\\
$|T_{+}|$ (entailed) & 248 & 154 \\
$|T_{-}|$ (contradictory) & 11 & 110 \\
$|T_{?}|$ (undetermined) & 1000 & 1000 \\
\midrule
\multicolumn{3}{l}{\emph{GCI1 evaluation split}}\\
$|T_{+}|$ (entailed) & 944 & 753 \\
$|T_{-}|$ (contradictory) & 45 & n/a \\
$|T_{?}|$ (undetermined) & 1000 & 997 \\
\midrule
\multicolumn{3}{l}{\emph{GCI2 evaluation split}}\\
$|T_{+}|$ (entailed) & 86 & 453 \\
$|T_{-}|$ (contradictory) & 527 & n/a \\
$|T_{?}|$ (undetermined) & 978 & 997 \\
\midrule
\multicolumn{3}{l}{\emph{GCI3 evaluation split}}\\
$|T_{+}|$ (entailed) & 180 & 12 \\
$|T_{-}|$ (contradictory) & 360 & n/a \\
$|T_{?}|$ (undetermined) & 988 & 994 \\
\bottomrule
\end{tabular}
\caption{Ontology sizes and evaluation-split composition. Axiom counts
exclude trivial axioms; ``n/a'' marks a quantity that is not applicable
(the GO-plus GCI1, GCI2, and GCI3 splits have no contradictory set,
whereas Pizza has one for every normal form).}
\label{app:tab-datastats}
\end{table}

\section{Full experimental results}
\label{app:results}

\begin{figure}[p]
\centering
\begin{subfigure}[b]{0.32\linewidth}
  \includegraphics[width=\linewidth]{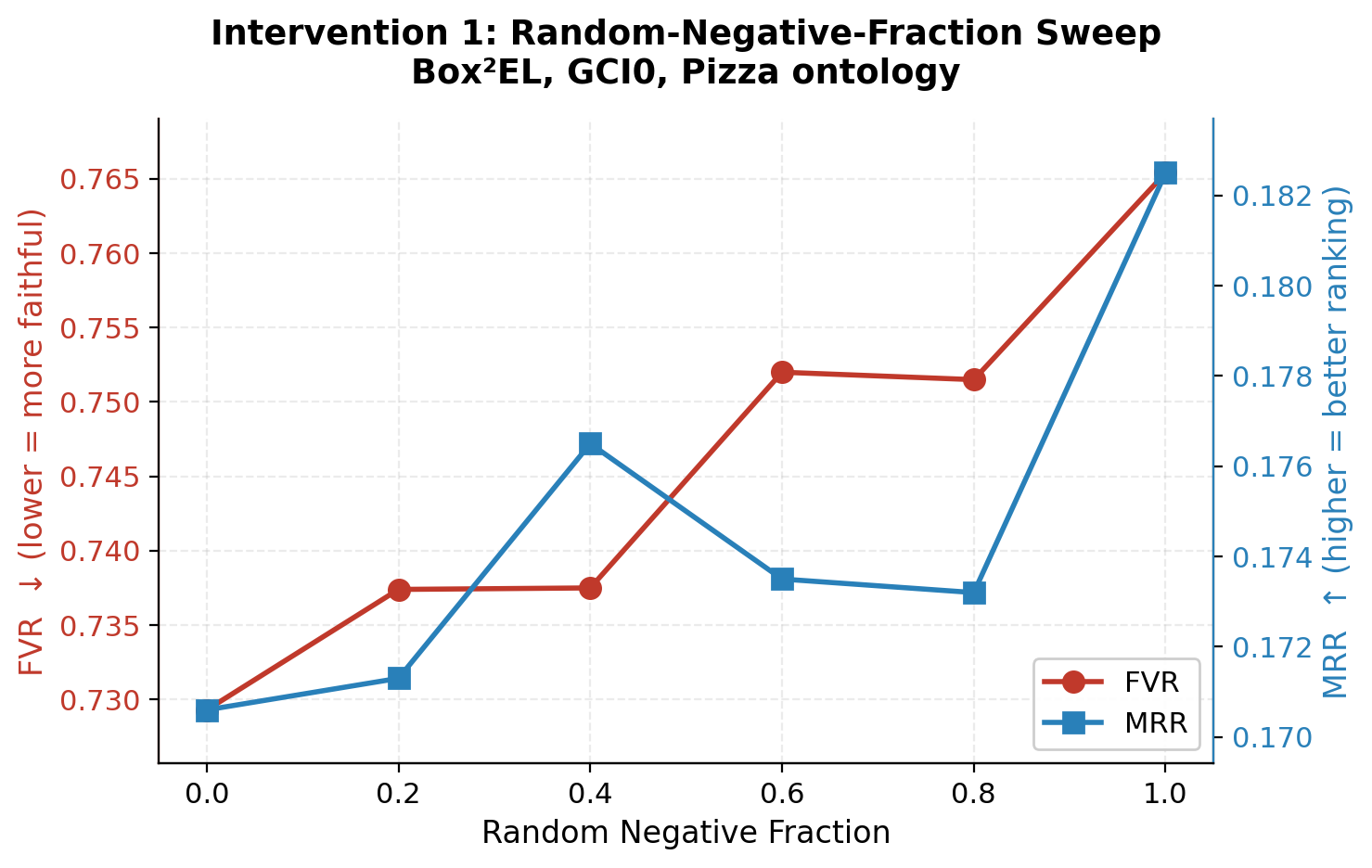}
  \caption{Box2EL, GCI0}
\end{subfigure}\hfill
\begin{subfigure}[b]{0.32\linewidth}
  \includegraphics[width=\linewidth]{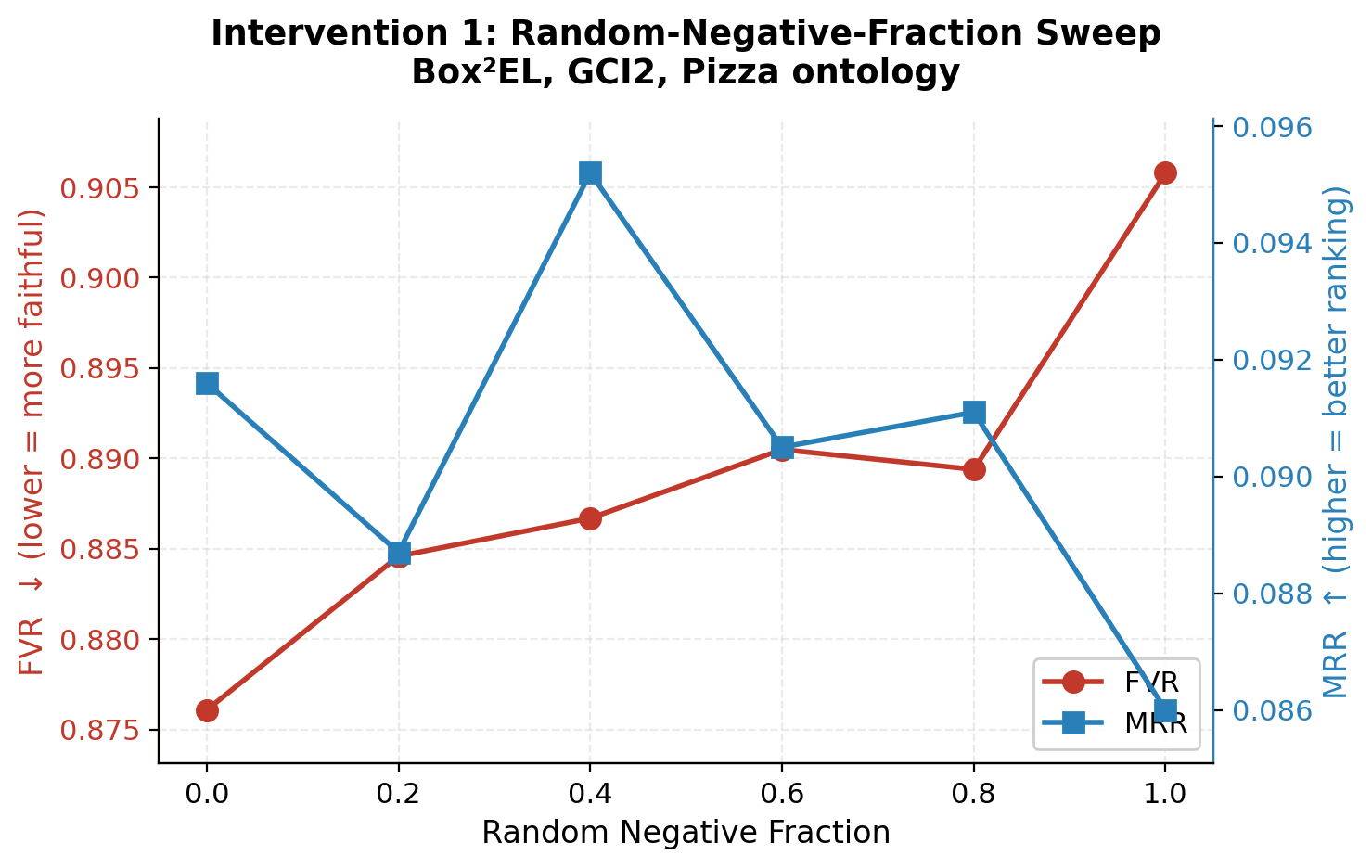}
  \caption{Box2EL, GCI2}
\end{subfigure}\hfill
\begin{subfigure}[b]{0.32\linewidth}
  \includegraphics[width=\linewidth]{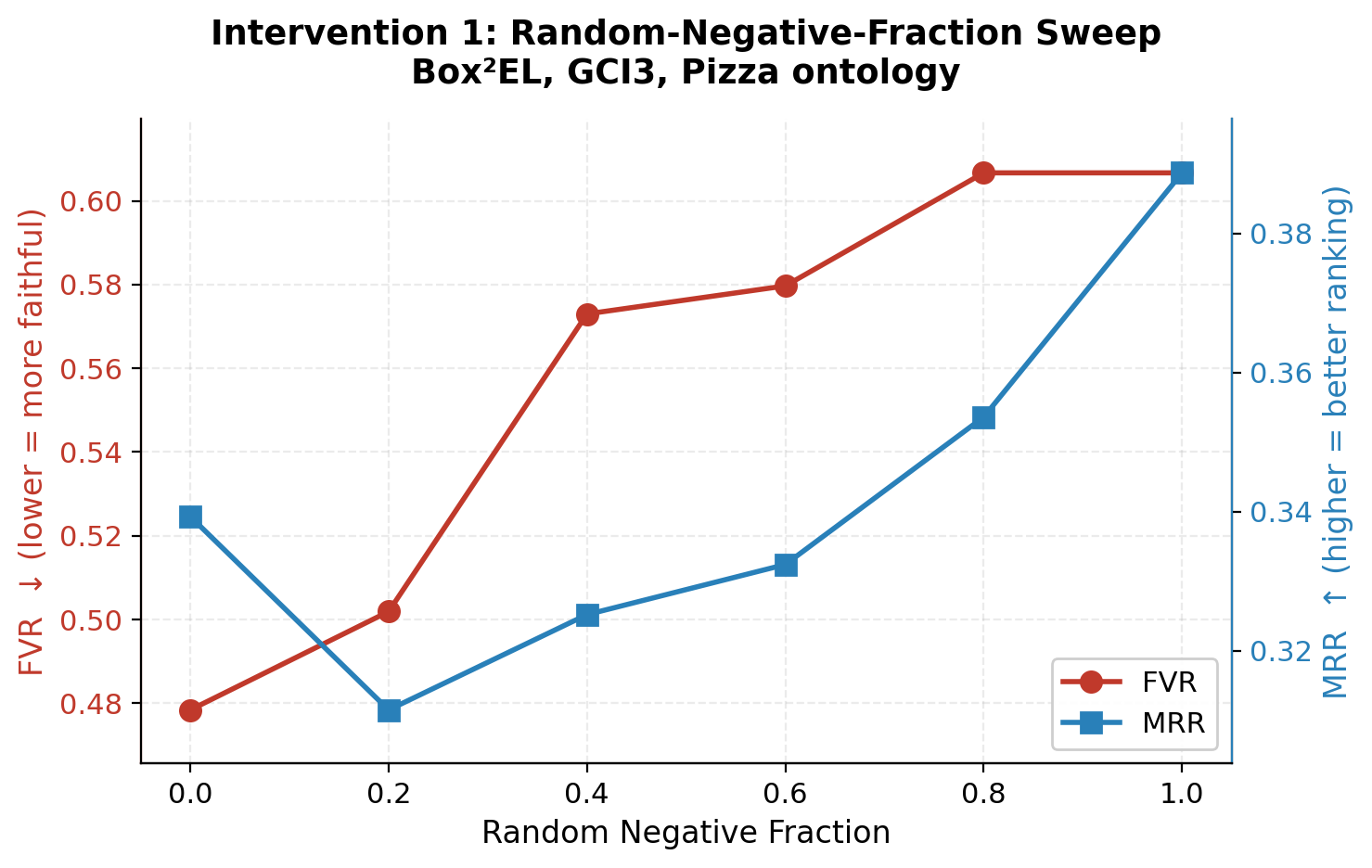}
  \caption{Box2EL, GCI3}
\end{subfigure}

\medskip
\begin{subfigure}[b]{0.32\linewidth}
  \includegraphics[width=\linewidth]{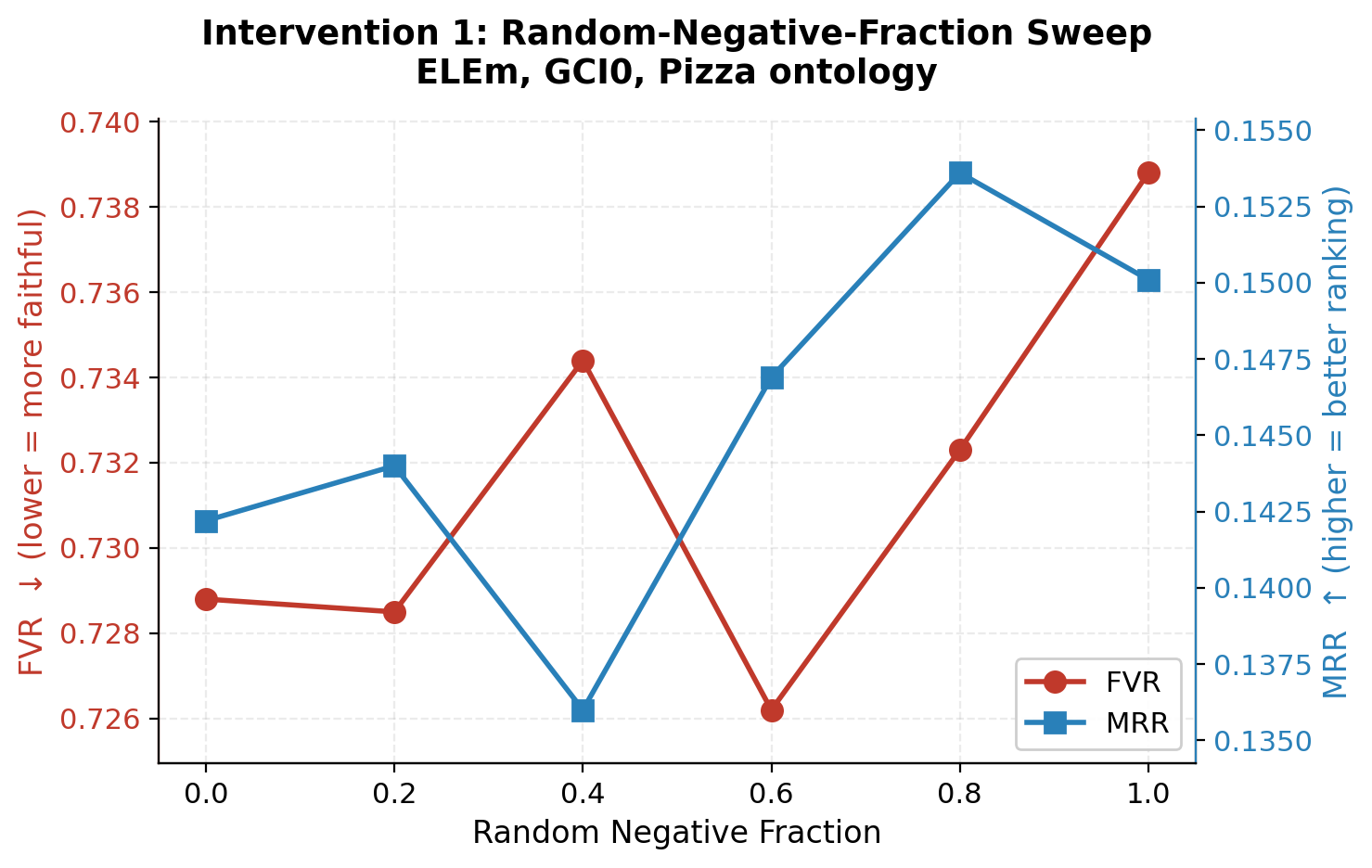}
  \caption{ELEmbeddings, GCI0}
\end{subfigure}\hfill
\begin{subfigure}[b]{0.32\linewidth}
  \includegraphics[width=\linewidth]{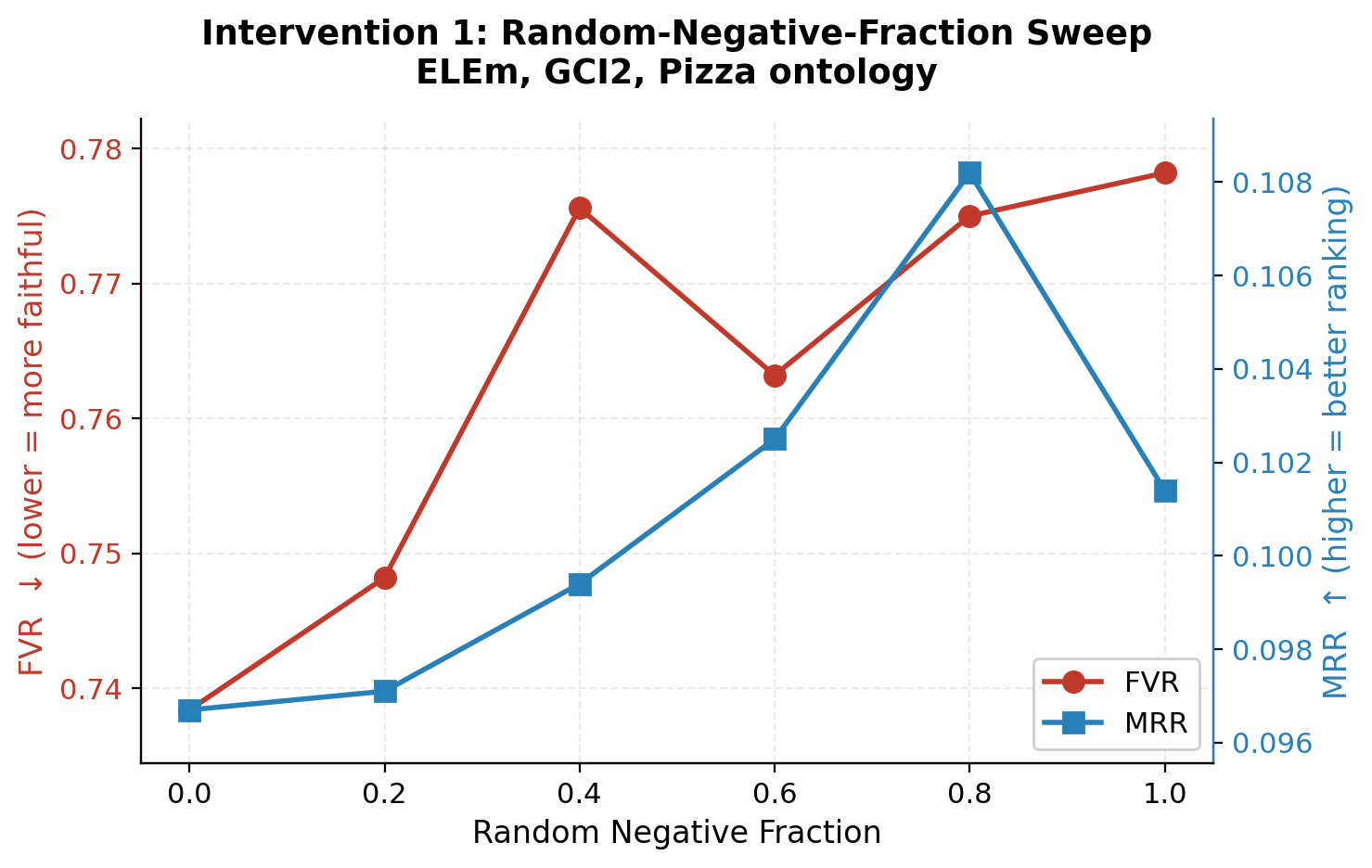}
  \caption{ELEmbeddings, GCI2}
\end{subfigure}\hfill
\begin{subfigure}[b]{0.32\linewidth}
  \includegraphics[width=\linewidth]{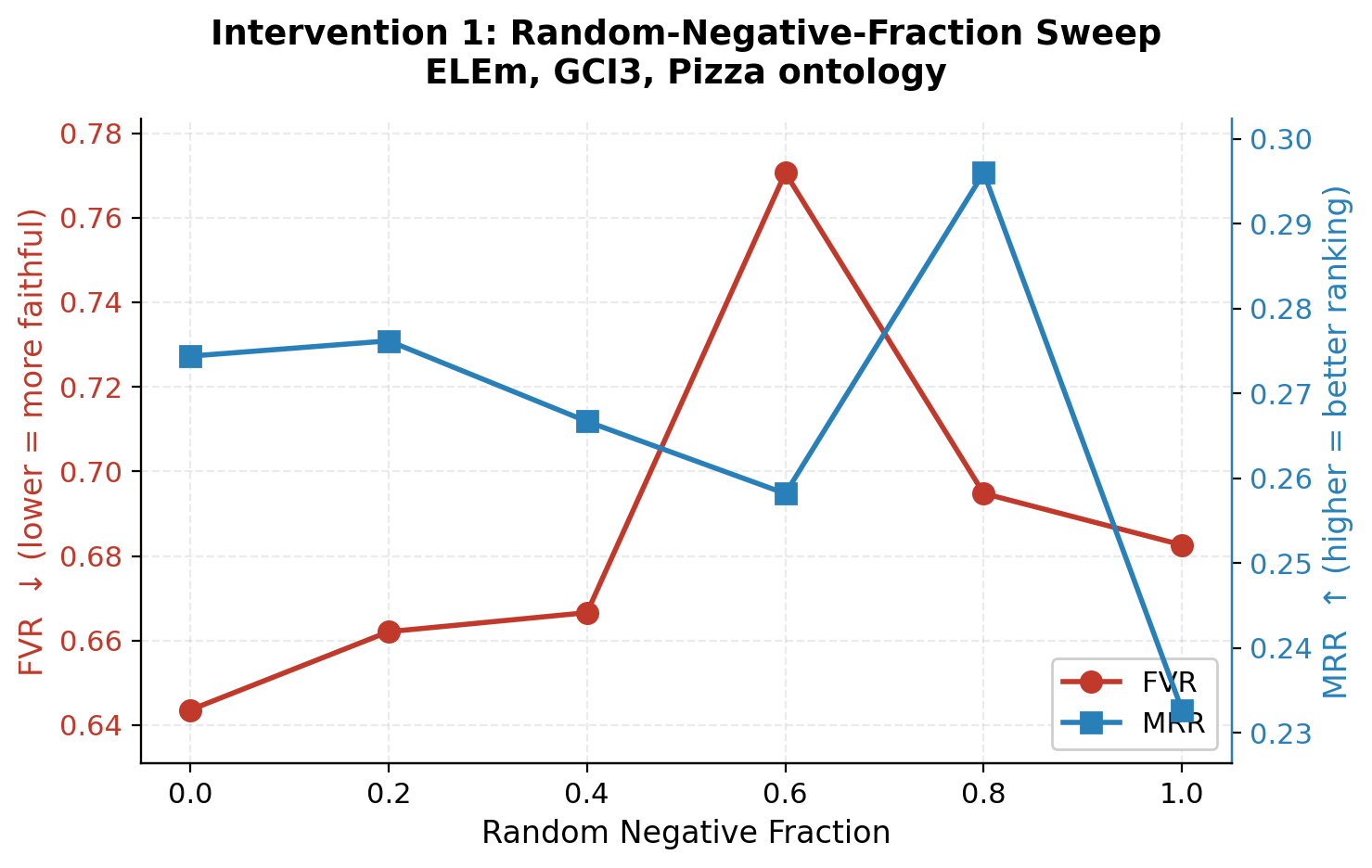}
  \caption{ELEmbeddings, GCI3}
\end{subfigure}

\medskip
\begin{subfigure}[b]{0.32\linewidth}
  \includegraphics[width=\linewidth]{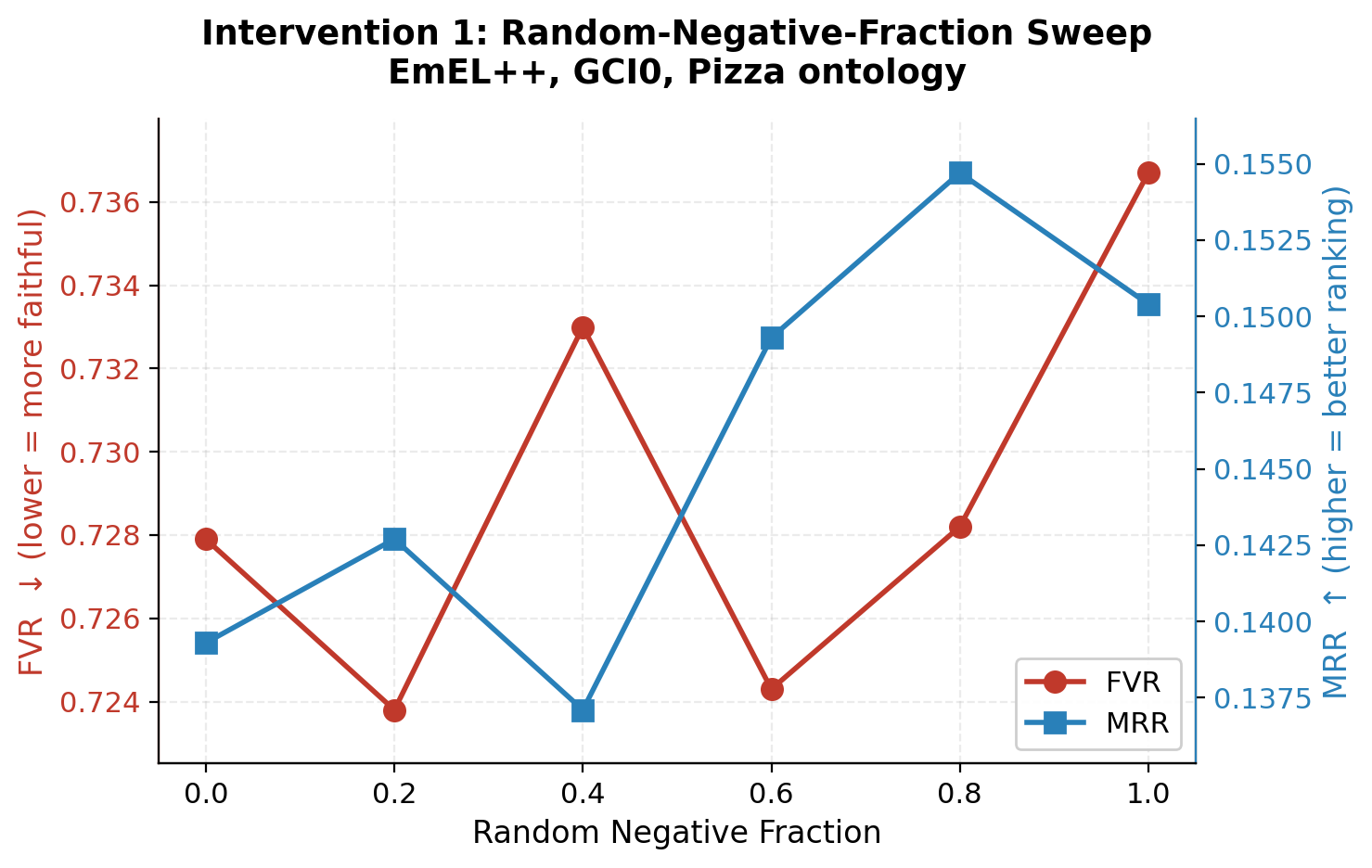}
  \caption{EmEL++, GCI0}
\end{subfigure}\hfill
\begin{subfigure}[b]{0.32\linewidth}
  \includegraphics[width=\linewidth]{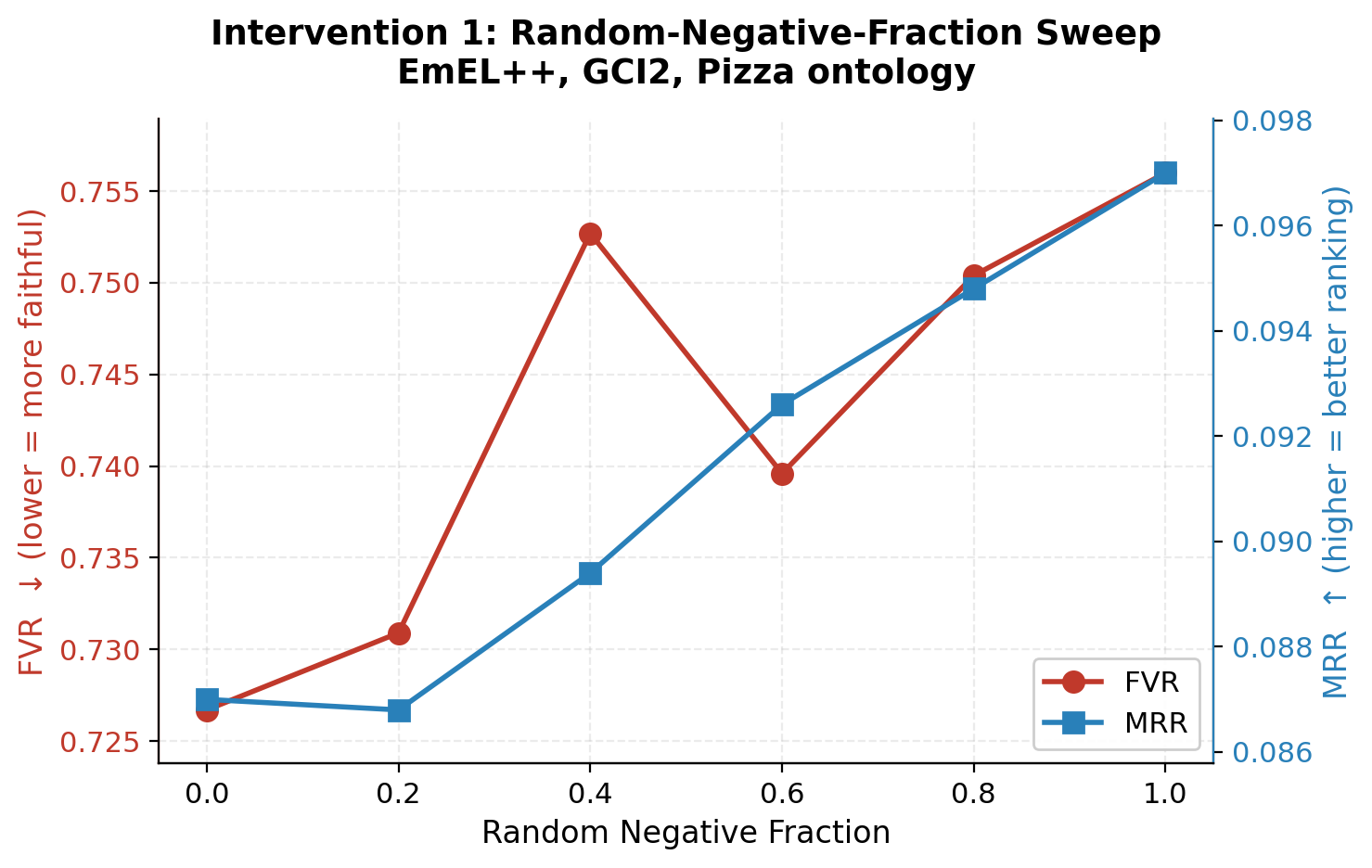}
  \caption{EmEL++, GCI2}
\end{subfigure}\hfill
\begin{subfigure}[b]{0.32\linewidth}
  \includegraphics[width=\linewidth]{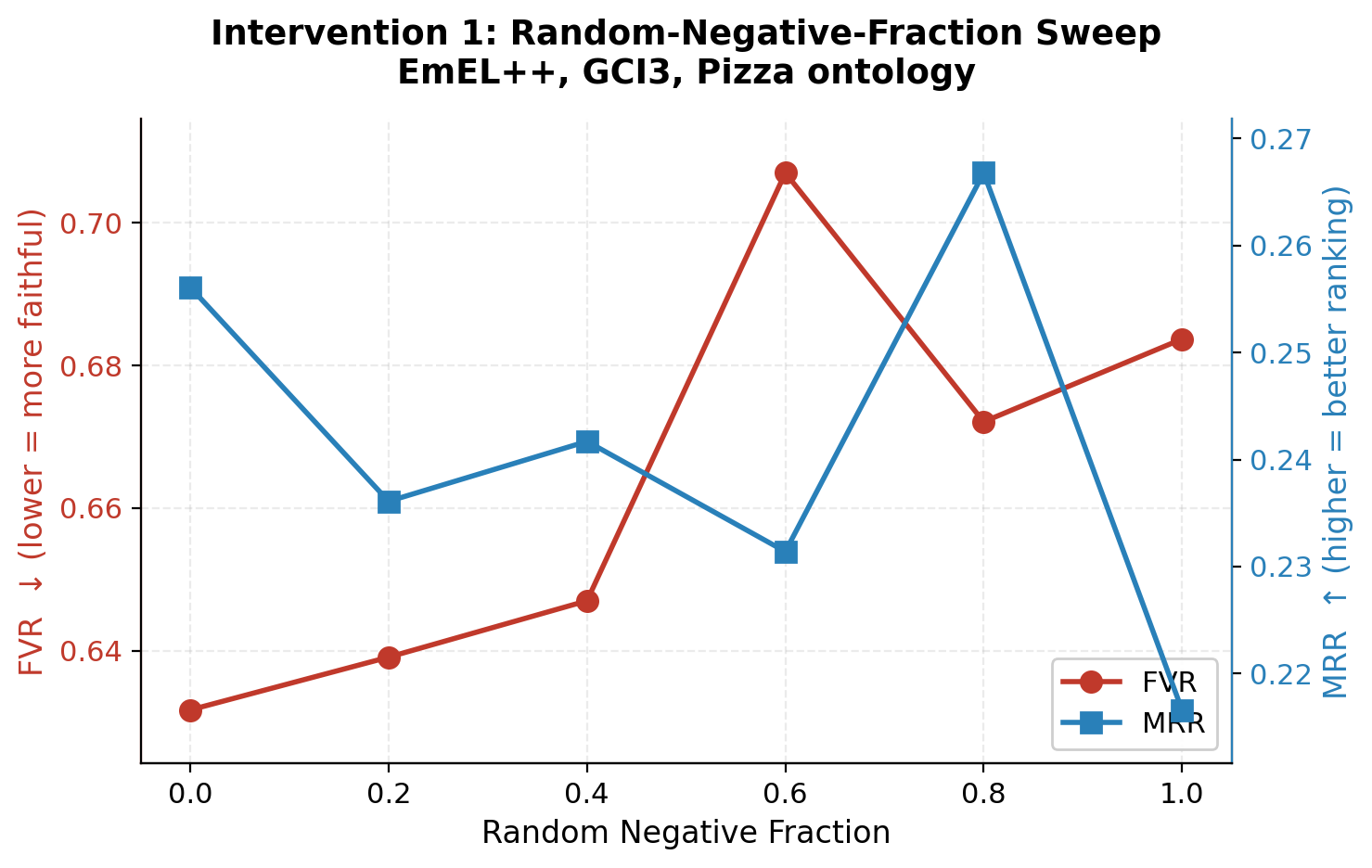}
  \caption{EmEL++, GCI3}
\end{subfigure}

\medskip
\begin{subfigure}[b]{0.32\linewidth}
  \includegraphics[width=\linewidth]{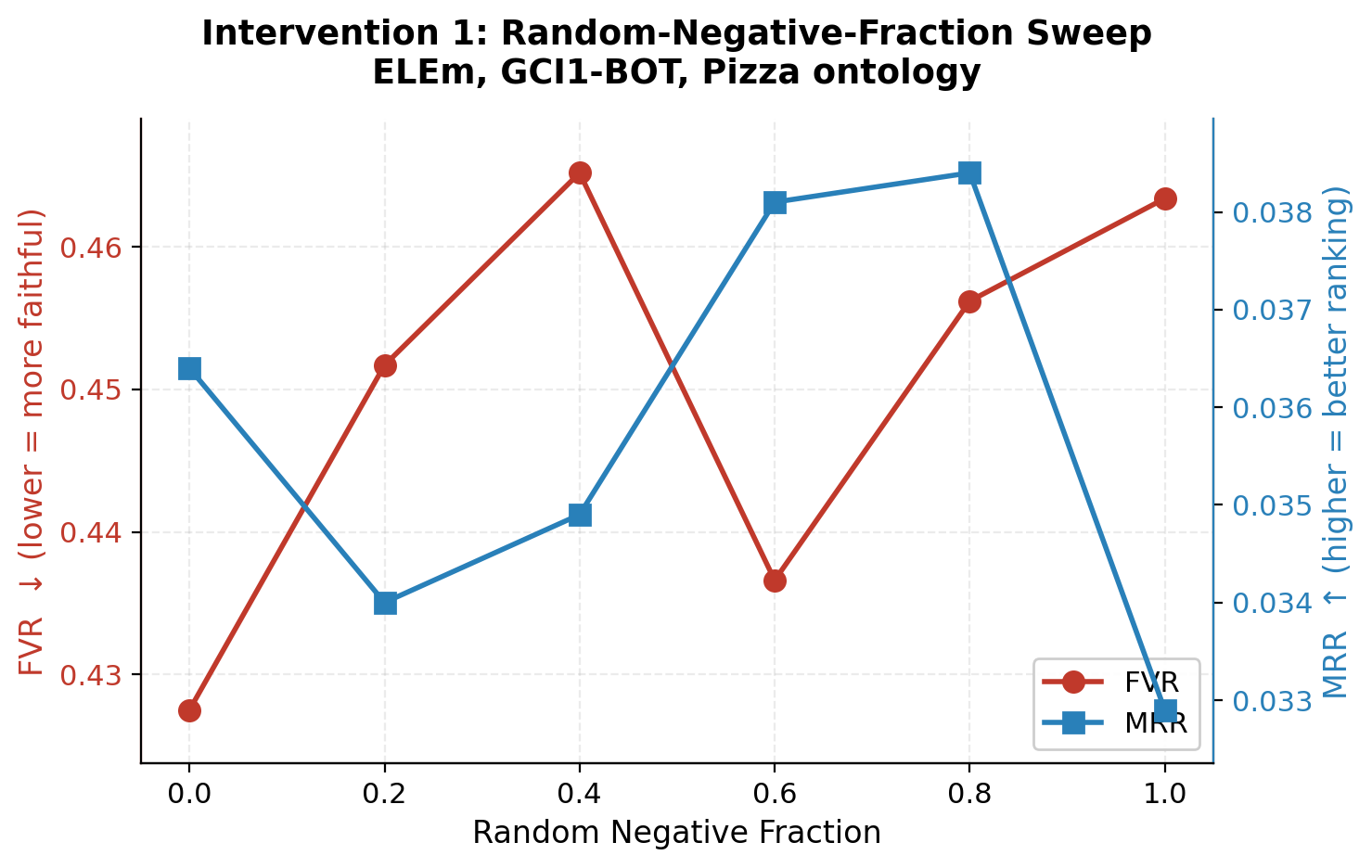}
  \caption{ELEmbeddings, GCI1$_\bot$}
\end{subfigure}\hfill
\begin{subfigure}[b]{0.32\linewidth}
  \includegraphics[width=\linewidth]{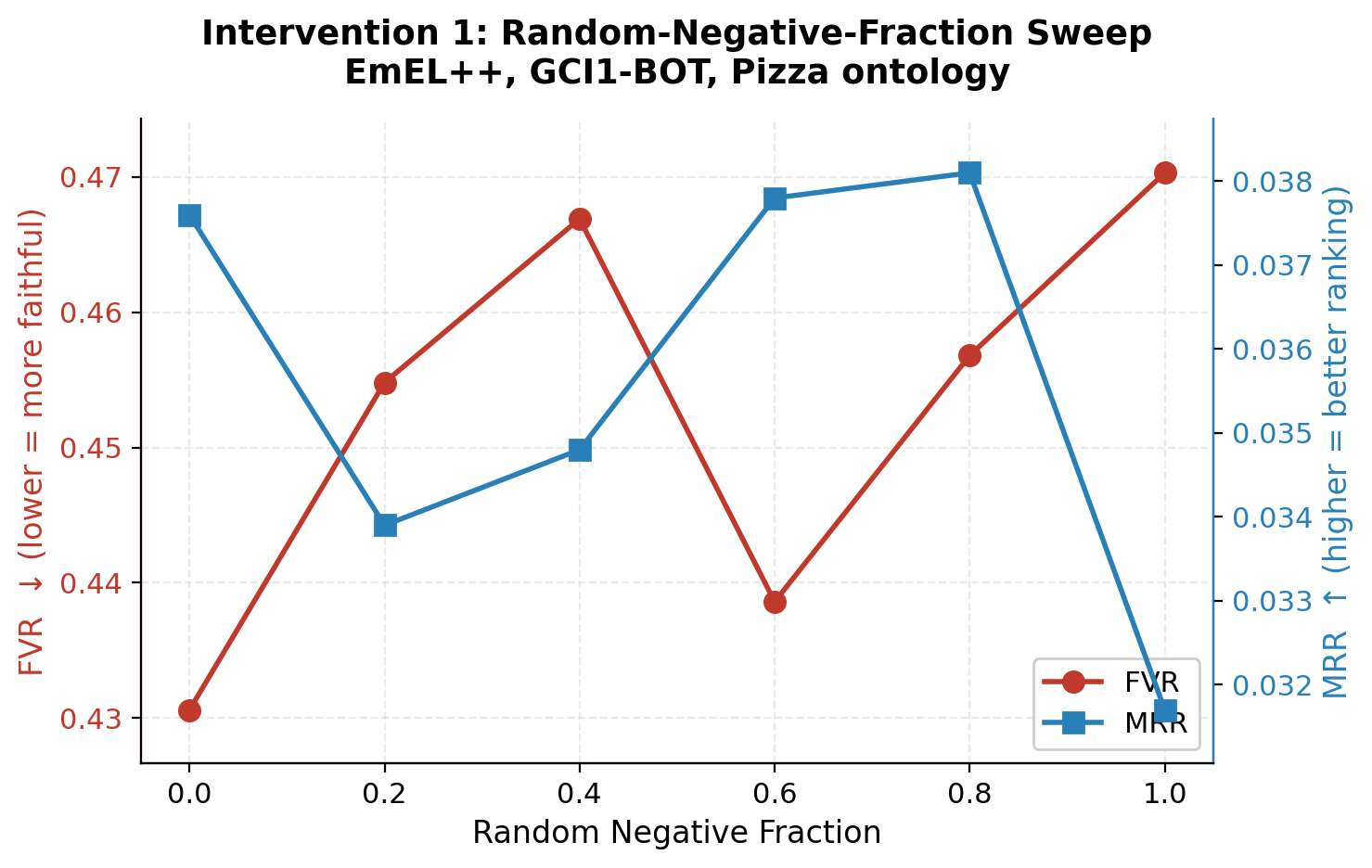}
  \caption{EmEL++, GCI1$_\bot$}
\end{subfigure}\hfill
\begin{subfigure}[b]{0.32\linewidth}
  \includegraphics[width=\linewidth]{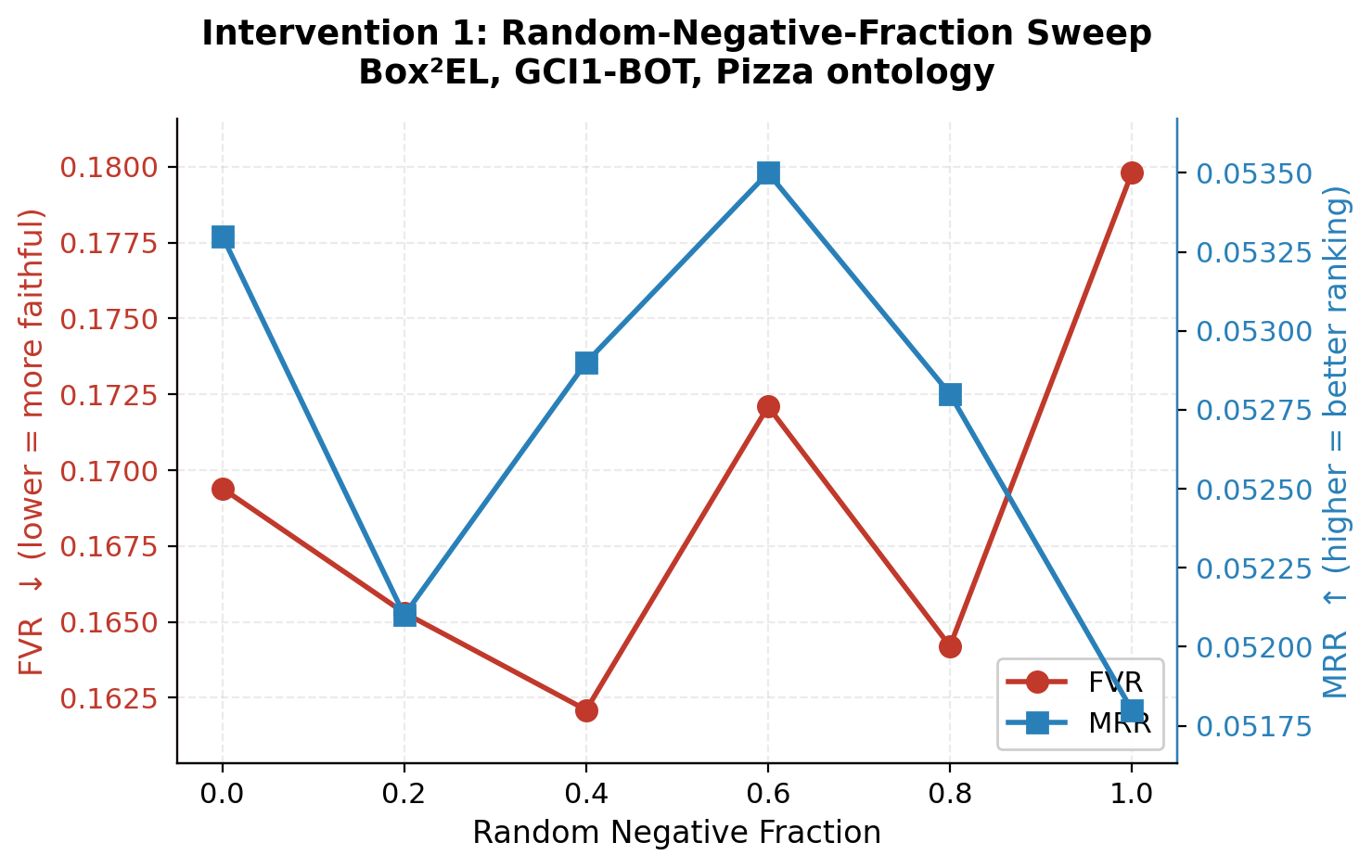}
  \caption{Box2EL, GCI1$_\bot$}
\end{subfigure}

\caption{Full negative-sampling sweep (Section~\ref{sec:training}) on
the Pizza ontology: FVR (left axis, lower is better) and MRR (right
axis) as the fraction of random-corruption negatives varies; the
remainder $p_\bot=1-x$ is drawn from $\Scon$. In all 12 panels the
all-contradiction endpoint ($x=0$) has lower FVR than the all-random
endpoint ($x=1$); the response is not monotone in between, and on GCI0
the ball-based models' effect is within single-seed noise.}
\label{fig:sweep-all}
\end{figure}

Tables~\ref{app:tab1} to~\ref{app:tab10} report all metrics
($\mathrm{AUC}_{+/-}$, Adm@$10$, FVR, Pair-FVR, $\rho_P$, MRR, Hits@10)
for every model, normal form, and dataset. The geometric baselines
(TransE, DistMult, ComplEx, BoxE) operate on atomic and existential
axioms and are therefore evaluated on GCI0 and GCI2; for the
remaining forms, and for the GO-plus slice normal forms without a
contradictory set, we report
$\rho_P$, MRR, and Hits@10. The summary Tables~\ref{tab:models}
and~\ref{tab:constructs} of Section~\ref{sec:results} are drawn from
these numbers. Table~\ref{tab:constructs} aggregates them by normal
form, and Figure~\ref{fig:sweep-all} reports the full negative-sampling
sweep of Section~\ref{sec:training}: all three logic-geometric models
on all four normal forms.

\begin{table}[h]
\centering
\begin{tabular}{llrrr}
\toprule
Normal form & Axiom & $\mathrm{AUC}_{+/-}$ $\uparrow$ & Pair-FVR $\downarrow$ & $\rho_P$ $\uparrow$ \\
\midrule
GCI0 & $C\sqsubseteq D$ & 0.72 & 0.53 & \phantom{-}0.59 \\
GCI1 & $C\sqcap D\sqsubseteq E$ & 0.15 & 0.80 & -0.27 \\
GCI2 & $C\sqsubseteq\exists r.D$ & 0.50 & 0.57 & \phantom{-}0.43 \\
GCI3 & $\exists r.C\sqsubseteq D$ & 0.84 & 0.55 & \phantom{-}0.22 \\
GCI1$_\bot$ & $C\sqcap D\sqsubseteq\bot$ & 0.97 & 0.06 & \phantom{-}0.42 \\
\bottomrule
\end{tabular}
\caption{Faithfulness is construct-dependent: mean over the
logic-geometric models (ELEmbeddings, EmEL++, Box2EL) by normal form, on
the Pizza ontology. Disjointness ($C\sqcap D\sqsubseteq\bot$) is the one
construct handled faithfully (high AUC, Pair-FVR near $0$); conjunctive
subsumption ($C\sqcap D\sqsubseteq E$) inverts (AUC below $0.5$, negative
$\rho_P$), driven by the ball models' loss
(Section~\ref{sec:results}; Appendix~\ref{app:elem}).}
\label{tab:constructs}
\end{table}

\begin{table}[]
    \centering
    \resizebox{\textwidth}{!}{%
    \begin{tabular}{c|c|c|c|c|c|c|c}
        Model name & $\mathrm{AUC}_{+/-}$ $\uparrow$ & Adm@10 $\uparrow$ & FVR $\downarrow$ & Pair-FVR $\downarrow$ & $\rho_P$ $\uparrow$ & MRR $\uparrow$ & Hits@10 $\uparrow$ \\
        \hline
         ELEm & 0.70 $\pm$ 0.01 & 0.98 $\pm$ 0.00 & 0.74 $\pm$ 0.00 & 0.48 $\pm$ 0.00 & 0.62 $\pm$ 0.02 & 0.14 $\pm$ 0.00 & 0.42 $\pm$ 0.01 \\
         \hline
         Box2EL & 0.74 $\pm$ 0.01 & 0.99 $\pm$ 0.00 & 0.76 $\pm$ 0.01 & 0.60 $\pm$ 0.01 & 0.53 $\pm$ 0.00 & 0.17 $\pm$ 0.01 & 0.51 $\pm$ 0.01 \\
         \hline
         EmEL++ & 0.73 $\pm$ 0.01 & 0.98 $\pm$ 0.01 & 0.73 $\pm$ 0.01 & 0.49 $\pm$ 0.00 & 0.63 $\pm$ 0.05 & 0.15 $\pm$ 0.01 & 0.41 $\pm$ 0.04 \\
         \hline
         BoxE & 0.42 $\pm$ 0.06 & 0.94 $\pm$ 0.01 & 0.90 $\pm$ 0.03 & 0.73 $\pm$ 0.02 & 0.16 $\pm$ 0.04 & 0.10 $\pm$ 0.00 & 0.23 $\pm$ 0.04 \\
         \hline
         ComplEx & 0.53 $\pm$ 0.03 & 0.94 $\pm$ 0.01 & 0.82 $\pm$ 0.01 & 0.45 $\pm$ 0.01 & -0.13 $\pm$ 0.03 & 0.08 $\pm$ 0.01 & 0.11 $\pm$ 0.01 \\
         \hline
         DistMult & 0.52 $\pm$ 0.09 & 0.94 $\pm$ 0.01 & 0.82 $\pm$ 0.05 & 0.47 $\pm$ 0.08 & -0.03 $\pm$ 0.08 & 0.05 $\pm$ 0.02 & 0.08 $\pm$ 0.06 \\
         \hline
         TransE & 0.61 $\pm$ 0.04 & 0.94 $\pm$ 0.01 & 0.79 $\pm$ 0.03 & 0.61 $\pm$ 0.05 & 0.18 $\pm$ 0.07 & 0.20 $\pm$ 0.00 & 0.32 $\pm$ 0.03 \\
    \end{tabular}}
    \caption{GCI0, Pizza}
    \label{app:tab1}
\end{table}

\begin{table}[]
    \centering
    \resizebox{\textwidth}{!}{%
    \begin{tabular}{c|c|c|c|c|c|c|c}
        Model name & $\mathrm{AUC}_{+/-}$ $\uparrow$ & Adm@10 $\uparrow$ & FVR $\downarrow$ & Pair-FVR $\downarrow$ & $\rho_P$ $\uparrow$ & MRR $\uparrow$ & Hits@10 $\uparrow$ \\
        \hline
         ELEm & 0.84 $\pm$ 0.01 & 1.00 $\pm$ 0.00 & 0.60 $\pm$ 0.01 & 0.47 $\pm$ 0.01 & 0.53 $\pm$ 0.01 & 0.16 $\pm$ 0.01 & 0.46 $\pm$ 0.01 \\
         \hline
         Box2EL & 0.82 $\pm$ 0.01 & 1.00 $\pm$ 0.00 & 0.50 $\pm$ 0.01 & 0.25 $\pm$ 0.02 & 0.63 $\pm$ 0.01 & 0.09 $\pm$ 0.00 & 0.26 $\pm$ 0.01 \\
         \hline
         EmEL++ & 0.83 $\pm$ 0.01 & 1.00 $\pm$ 0.00 & 0.62 $\pm$ 0.02 & 0.49 $\pm$ 0.02 & 0.57 $\pm$ 0.01 & 0.15 $\pm$ 0.01 & 0.45 $\pm$ 0.02 \\
         \hline
         BoxE & 0.61 $\pm$ 0.01 & 1.00 $\pm$ 0.00 & 0.74 $\pm$ 0.01 & 0.33 $\pm$ 0.02 & 0.24 $\pm$ 0.02 & 0.16 $\pm$ 0.00 & 0.20 $\pm$ 0.00 \\
         \hline
         ComplEx & 0.55 $\pm$ 0.05 & 1.00 $\pm$ 0.00 & 0.78 $\pm$ 0.04 & 0.49 $\pm$ 0.08 & 0.01 $\pm$ 0.05 & 0.16 $\pm$ 0.00 & 0.19 $\pm$ 0.00 \\
         \hline
         DistMult & 0.27 $\pm$ 0.07 & 1.00 $\pm$ 0.00 & 0.95 $\pm$ 0.03 & 0.76 $\pm$ 0.13 & 0.01 $\pm$ 0.01 & 0.04 $\pm$ 0.02 & 0.09 $\pm$ 0.02 \\
         \hline
         TransE & 0.61 $\pm$ 0.06 & 1.00 $\pm$ 0.00 & 0.73 $\pm$ 0.02 & 0.48 $\pm$ 0.06 & 0.34 $\pm$ 0.05 & 0.24 $\pm$ 0.01 & 0.36 $\pm$ 0.00 \\
    \end{tabular}}
    \caption{GCI0, GO-plus slice}
    \label{app:tab2}
\end{table}

\begin{table}[]
    \centering
    \resizebox{\textwidth}{!}{%
    \begin{tabular}{c|c|c|c|c|c|c|c}
        Model name & $\mathrm{AUC}_{+/-}$ $\uparrow$ & Adm@10 $\uparrow$ & FVR $\downarrow$ & Pair-FVR $\downarrow$ & $\rho_P$ $\uparrow$ & MRR $\uparrow$ & Hits@10 $\uparrow$ \\
        \hline
         ELEm & 0.57 $\pm$ 0.01 & 1.00 $\pm$ 0.00 & 0.78 $\pm$ 0.00 & 0.52 $\pm$ 0.01 & 0.51 $\pm$ 0.01 & 0.10 $\pm$ 0.00 & 0.29 $\pm$ 0.01 \\
         \hline
         Box2EL & 0.35 $\pm$ 0.02 & 1.00 $\pm$ 0.00 & 0.90 $\pm$ 0.01 & 0.73 $\pm$ 0.02 & 0.22 $\pm$ 0.01 & 0.09 $\pm$ 0.00 & 0.18 $\pm$ 0.01 \\
         \hline
         EmEL++ & 0.59 $\pm$ 0.01 & 1.00 $\pm$ 0.00 & 0.76 $\pm$ 0.00 & 0.48 $\pm$ 0.01 & 0.55 $\pm$ 0.01 & 0.10 $\pm$ 0.00 & 0.29 $\pm$ 0.01 \\
         \hline
         BoxE & 0.34 $\pm$ 0.01 & 1.00 $\pm$ 0.00 & 0.91 $\pm$ 0.00 & 0.75 $\pm$ 0.01 & 0.22 $\pm$ 0.03 & 0.14 $\pm$ 0.01 & 0.20 $\pm$ 0.01 \\
         \hline
         ComplEx & 0.51 $\pm$ 0.03 & 1.00 $\pm$ 0.00 & 0.83 $\pm$ 0.01 & 0.51 $\pm$ 0.03 & -0.05 $\pm$ 0.01 & 0.11 $\pm$ 0.01 & 0.19 $\pm$ 0.02 \\
         \hline
         DistMult & 0.47 $\pm$ 0.02 & 1.00 $\pm$ 0.00 & 0.86 $\pm$ 0.01 & 0.48 $\pm$ 0.06 & -0.04 $\pm$ 0.03 & 0.09 $\pm$ 0.01 & 0.18 $\pm$ 0.02 \\
         \hline
         TransE & 0.45 $\pm$ 0.04 & 1.00 $\pm$ 0.00 & 0.88 $\pm$ 0.02 & 0.69 $\pm$ 0.03 & 0.25 $\pm$ 0.09 & 0.12 $\pm$ 0.01 & 0.22 $\pm$ 0.03 \\
    \end{tabular}}
    \caption{GCI2, Pizza}
    \label{app:tab3}
\end{table}

\begin{table}[]
    \centering
    \begin{tabular}{c|c|c|c}
        Model name & $\rho_P$ $\uparrow$ & MRR $\uparrow$ & Hits@10 $\uparrow$ \\
        \hline
         ELEm & 0.43 $\pm$ 0.02 & 0.06 $\pm$ 0.00 & 0.22 $\pm$ 0.00 \\
         \hline
         Box2EL & 0.30 $\pm$ 0.00 & 0.01 $\pm$ 0.00 & 0.01 $\pm$ 0.01 \\
         \hline
         EmEL++ & 0.49 $\pm$ 0.02 & 0.06 $\pm$ 0.00 & 0.22 $\pm$ 0.02 \\
         \hline
         BoxE & 0.15 $\pm$ 0.04 & 0.06 $\pm$ 0.00 & 0.07 $\pm$ 0.00 \\
         \hline
         ComplEx & -0.15 $\pm$ 0.05 & 0.06 $\pm$ 0.00 & 0.07 $\pm$ 0.00 \\
         \hline
         DistMult & -0.22 $\pm$ 0.14 & 0.01 $\pm$ 0.00 & 0.01 $\pm$ 0.01 \\
         \hline
         TransE & 0.28 $\pm$ 0.03 & 0.04 $\pm$ 0.00 & 0.09 $\pm$ 0.01 \\
    \end{tabular}
    \caption{GCI2, GO-plus slice}
    \label{app:tab4}
\end{table}

\begin{table}[]
    \centering
    \resizebox{\textwidth}{!}{%
    \begin{tabular}{c|c|c|c|c|c|c|c}
        Model name & $\mathrm{AUC}_{+/-}$ $\uparrow$ & Adm@10 $\uparrow$ & FVR $\downarrow$ & Pair-FVR $\downarrow$ & $\rho_P$ $\uparrow$ & MRR $\uparrow$ & Hits@10 $\uparrow$ \\
        \hline
         ELEm & 0.02 $\pm$ 0.00 & 1.00 $\pm$ 0.00 & 1.00 $\pm$ 0.00 & 0.94 $\pm$ 0.01 & -0.34 $\pm$ 0.01 & 0.06 $\pm$ 0.00 & 0.11 $\pm$ 0.01 \\
         \hline
         Box2EL & 0.39 $\pm$ 0.01 & 1.00 $\pm$ 0.00 & 0.91 $\pm$ 0.00 & 0.55 $\pm$ 0.01 & -0.09 $\pm$ 0.02 & 0.05 $\pm$ 0.00 & 0.13  $\pm$ 0.01 \\
         \hline
         EmEL++ & 0.05 $\pm$ 0.01 & 1.00 $\pm$ 0.00 & 1.00 $\pm$ 0.00 & 0.91 $\pm$ 0.02 & -0.31 $\pm$ 0.03 & 0.06 $\pm$ 0.00 & 0.11 $\pm$ 0.00 \\
    \end{tabular}}
    \caption{GCI1, Pizza}
    \label{app:tab5}
\end{table}

\begin{table}[]
    \centering
    \begin{tabular}{c|c|c|c}
        Model name & $\rho_P$ $\uparrow$ & MRR $\uparrow$ & Hits@10 $\uparrow$ \\
        \hline
         ELEm & -0.39 $\pm$ 0.02 & 0.08 $\pm$ 0.00 & 0.20 $\pm$ 0.01 \\
         \hline
         Box2EL & 0.12 $\pm$ 0.01 & 0.04 $\pm$ 0.00 & 0.06 $\pm$ 0.01 \\
         \hline
         EmEL++ & -0.40 $\pm$ 0.01 & 0.09 $\pm$ 0.00 & 0.20 $\pm$ 0.01 \\
    \end{tabular}
    \caption{GCI1, GO-plus slice}
    \label{app:tab6}
\end{table}

\begin{table}[]
    \centering
    \resizebox{\textwidth}{!}{%
    \begin{tabular}{c|c|c|c|c|c|c|c}
        Model name & $\mathrm{AUC}_{+/-}$ $\uparrow$ & Adm@10 $\uparrow$ & FVR $\downarrow$ & Pair-FVR $\downarrow$ & $\rho_P$ $\uparrow$ & MRR $\uparrow$ & Hits@10 $\uparrow$ \\
        \hline
         ELEm & 0.96 $\pm$ 0.00 & 1.00 $\pm$ 0.00 & 0.47 $\pm$ 0.01 & 0.08 $\pm$ 0.01 & 0.43 $\pm$ 0.00 & 0.03 $\pm$ 0.00 & 0.02 $\pm$ 0.00 \\
         \hline
         Box2EL & 1.00 $\pm$ 0.00 & 1.00 $\pm$ 0.00 & 0.17 $\pm$ 0.01 & 0.01 $\pm$ 0.00 & 0.41 $\pm$ 0.00 & 0.05 $\pm$ 0.00 & 0.01 $\pm$ 0.00 \\
         \hline
         EmEL++ & 0.96 $\pm$ 0.00 & 1.00 $\pm$ 0.00 & 0.47 $\pm$ 0.01 & 0.07 $\pm$ 0.01 & 0.43 $\pm$ 0.00 & 0.03 $\pm$ 0.00 & 0.02 $\pm$ 0.00 \\
    \end{tabular}}
    \caption{GCI1$_\bot$, Pizza}
    \label{app:tab7}
\end{table}

\begin{table}[]
    \centering
    \resizebox{\textwidth}{!}{%
    \begin{tabular}{c|c|c|c|c|c|c|c}
        Model name & $\mathrm{AUC}_{+/-}$ $\uparrow$ & Adm@10 $\uparrow$ & FVR $\downarrow$ & Pair-FVR $\downarrow$ & $\rho_P$ $\uparrow$ & MRR $\uparrow$ & Hits@10 $\uparrow$ \\
        \hline
         ELEm & 0.95 $\pm$ 0.01 & 1.00 $\pm$ 0.00 & 0.65 $\pm$ 0.01 & 0.06 $\pm$ 0.00 & 0.61 $\pm$ 0.01 & 0.01 $\pm$ 0.00 & 0.03 $\pm$ 0.01 \\
         \hline
         Box2EL & 0.99 $\pm$ 0.00 & 1.00 $\pm$ 0.00 & 0.50 $\pm$ 0.01 & 0.03 $\pm$ 0.00 & 0.73 $\pm$ 0.00 & 0.01 $\pm$ 0.00 & 0.04 $\pm$ 0.00 \\
         \hline
         EmEL++ & 0.95 $\pm$ 0.01 & 1.00 $\pm$ 0.00 & 0.65 $\pm$ 0.01 & 0.06 $\pm$ 0.00 & 0.61 $\pm$ 0.01 & 0.01 $\pm$ 0.00 & 0.03 $\pm$ 0.01 \\
    \end{tabular}}
    \caption{GCI1$_\bot$, GO-plus slice}
    \label{app:tab8}
\end{table}

\begin{table}[]
    \centering
    \resizebox{\textwidth}{!}{%
    \begin{tabular}{c|c|c|c|c|c|c|c}
        Model name & $\mathrm{AUC}_{+/-}$ $\uparrow$ & Adm@10 $\uparrow$ & FVR $\downarrow$ & Pair-FVR $\downarrow$ & $\rho_P$ $\uparrow$ & MRR $\uparrow$ & Hits@10 $\uparrow$ \\
        \hline
         ELEm & 0.75 $\pm$ 0.01 & 1.00 $\pm$ 0.00 & 0.71 $\pm$ 0.02 & 0.56 $\pm$ 0.03 & 0.26 $\pm$ 0.01 & 0.23 $\pm$ 0.03 & 0.65 $\pm$ 0.01 \\
         \hline
         Box2EL & 0.97 $\pm$ 0.00 & 1.00 $\pm$ 0.00 & 0.62 $\pm$ 0.01 & 0.55 $\pm$ 0.01 & 0.09 $\pm$ 0.01 & 0.39 $\pm$ 0.02 & 0.77 $\pm$ 0.02 \\
         \hline
         EmEL++ & 0.79 $\pm$ 0.02 & 1.00 $\pm$ 0.00 & 0.68 $\pm$ 0.01 & 0.54 $\pm$ 0.01 & 0.31 $\pm$ 0.01 & 0.22 $\pm$ 0.02 & 0.60 $\pm$ 0.03 \\
    \end{tabular}}
    \caption{GCI3, Pizza}
    \label{app:tab9}
\end{table}

\begin{table}[]
    \centering
    \begin{tabular}{c|c|c|c}
        Model name & $\rho_P$ $\uparrow$ & MRR $\uparrow$ & Hits@10 $\uparrow$ \\
        \hline
         ELEm & 0.55 $\pm$ 0.01 & 0.01 $\pm$ 0.00 & 0.01 $\pm$ 0.00 \\
         \hline
         Box2EL & 0.42 $\pm$ 0.01 & 0.01 $\pm$ 0.00 & 0.01 $\pm$ 0.00 \\
         \hline
         EmEL++ & 0.53 $\pm$ 0.00 & 0.01 $\pm$ 0.00 & 0.01 $\pm$ 0.00 \\
    \end{tabular}
    \caption{GCI3, GO-plus slice}
    \label{app:tab10}
\end{table}

\subsection*{Multi-seed intervention-1 sweep}
\label{app:sweep-multiseed}

Tables~\ref{app:tab-sweep-b2el-pizza-gci0}
onwards report the mean and standard deviation across five seeds of the
intervention-1 sweep of Section~\ref{sec:training}, for all three
logic-geometric models on the four normal forms of
Figure~\ref{fig:sweep-all} (atomic subsumption GCI0, disjointness
GCI1$_\bot$, and the existential forms GCI2 and GCI3). Each table
varies $x$, the fraction of random-corruption negatives, with the
remaining $1-x$ drawn from $\Scon$. Across all twelve panels the
all-contradiction endpoint ($x=0$) attains lower mean FVR than the
all-random endpoint ($x=1$); the effect is largest on Box2EL/GCI3 (a
difference of $0.13$ in mean FVR) and smallest on the disjointness
form, where for Box2EL FVR is already close to its floor. Consistent
with Table~\ref{tab:models}, MRR does not track FVR across the sweep,
reproducing the rank--faithfulness dissociation as an intervention
within a single model.

\begin{table}[h]
\centering
\begin{tabular}{crr}
\toprule
$x$ & FVR $\downarrow$ & MRR $\uparrow$ \\
\midrule
0.0 & $0.7303 \pm 0.0087$ & $0.1710 \pm 0.0076$ \\
0.2 & $0.7328 \pm 0.0122$ & $0.1701 \pm 0.0093$ \\
0.4 & $0.7444 \pm 0.0054$ & $0.1721 \pm 0.0067$ \\
0.6 & $0.7516 \pm 0.0050$ & $0.1743 \pm 0.0066$ \\
0.8 & $0.7524 \pm 0.0082$ & $0.1728 \pm 0.0048$ \\
1.0 & $0.7493 \pm 0.0114$ & $0.1748 \pm 0.0057$ \\
\bottomrule
\end{tabular}
\caption{Multi-seed (5 seeds) sweep for Box2EL, Pizza ontology, GCI0.
$x$ is the fraction of random-corruption negatives, with the remaining
$1-x$ drawn from $\Scon$ (matching Figure~\ref{fig:sweep-all}). Values
are mean $\pm$ standard deviation.}
\label{app:tab-sweep-b2el-pizza-gci0}
\end{table}

\begin{table}[h]
\centering
\begin{tabular}{crr}
\toprule
$x$ & FVR $\downarrow$ & MRR $\uparrow$ \\
\midrule
0.0 & $0.7335 \pm 0.0035$ & $0.1401 \pm 0.0024$ \\
0.2 & $0.7377 \pm 0.0063$ & $0.1426 \pm 0.0030$ \\
0.4 & $0.7344 \pm 0.0051$ & $0.1439 \pm 0.0057$ \\
0.6 & $0.7383 \pm 0.0078$ & $0.1431 \pm 0.0049$ \\
0.8 & $0.7395 \pm 0.0069$ & $0.1470 \pm 0.0083$ \\
1.0 & $0.7431 \pm 0.0052$ & $0.1452 \pm 0.0041$ \\
\bottomrule
\end{tabular}
\caption{Multi-seed (5 seeds) sweep for ELEmbeddings, Pizza ontology,
GCI0; conventions as in Table~\ref{app:tab-sweep-b2el-pizza-gci0}.}
\label{app:tab-sweep-elem-pizza-gci0}
\end{table}

\begin{table}[h]
\centering
\begin{tabular}{crr}
\toprule
$x$ & FVR $\downarrow$ & MRR $\uparrow$ \\
\midrule
0.0 & $0.7304 \pm 0.0038$ & $0.1409 \pm 0.0062$ \\
0.2 & $0.7357 \pm 0.0068$ & $0.1402 \pm 0.0025$ \\
0.4 & $0.7340 \pm 0.0023$ & $0.1461 \pm 0.0069$ \\
0.6 & $0.7354 \pm 0.0074$ & $0.1444 \pm 0.0049$ \\
0.8 & $0.7357 \pm 0.0077$ & $0.1485 \pm 0.0082$ \\
1.0 & $0.7398 \pm 0.0032$ & $0.1458 \pm 0.0038$ \\
\bottomrule
\end{tabular}
\caption{Multi-seed (5 seeds) sweep for EmEL++, Pizza ontology, GCI0;
conventions as in Table~\ref{app:tab-sweep-b2el-pizza-gci0}.}
\label{app:tab-sweep-emel-pizza-gci0}
\end{table}

\begin{table}[h]
\centering
\begin{tabular}{crr}
\toprule
$x$ & FVR $\downarrow$ & MRR $\uparrow$ \\
\midrule
0.0 & $0.1695 \pm 0.0042$ & $0.0537 \pm 0.0013$ \\
0.2 & $0.1707 \pm 0.0051$ & $0.0517 \pm 0.0003$ \\
0.4 & $0.1725 \pm 0.0087$ & $0.0515 \pm 0.0013$ \\
0.6 & $0.1694 \pm 0.0070$ & $0.0526 \pm 0.0012$ \\
0.8 & $0.1746 \pm 0.0075$ & $0.0517 \pm 0.0020$ \\
1.0 & $0.1789 \pm 0.0066$ & $0.0516 \pm 0.0020$ \\
\bottomrule
\end{tabular}
\caption{Multi-seed (5 seeds) sweep for Box2EL, Pizza ontology,
GCI1$_\bot$ (disjointness); conventions as in
Table~\ref{app:tab-sweep-b2el-pizza-gci0}.}
\label{app:tab-sweep-b2el-pizza-gci1bot}
\end{table}

\begin{table}[h]
\centering
\begin{tabular}{crr}
\toprule
$x$ & FVR $\downarrow$ & MRR $\uparrow$ \\
\midrule
0.0 & $0.4323 \pm 0.0102$ & $0.0351 \pm 0.0008$ \\
0.2 & $0.4452 \pm 0.0053$ & $0.0341 \pm 0.0020$ \\
0.4 & $0.4484 \pm 0.0204$ & $0.0343 \pm 0.0022$ \\
0.6 & $0.4452 \pm 0.0251$ & $0.0340 \pm 0.0024$ \\
0.8 & $0.4563 \pm 0.0178$ & $0.0335 \pm 0.0035$ \\
1.0 & $0.4580 \pm 0.0106$ & $0.0325 \pm 0.0009$ \\
\bottomrule
\end{tabular}
\caption{Multi-seed (5 seeds) sweep for ELEmbeddings, Pizza ontology,
GCI1$_\bot$ (disjointness); conventions as in
Table~\ref{app:tab-sweep-b2el-pizza-gci0}.}
\label{app:tab-sweep-elem-pizza-gci1bot}
\end{table}

\begin{table}[h]
\centering
\begin{tabular}{crr}
\toprule
$x$ & FVR $\downarrow$ & MRR $\uparrow$ \\
\midrule
0.0 & $0.4330 \pm 0.0079$ & $0.0351 \pm 0.0014$ \\
0.2 & $0.4475 \pm 0.0070$ & $0.0339 \pm 0.0019$ \\
0.4 & $0.4516 \pm 0.0184$ & $0.0339 \pm 0.0012$ \\
0.6 & $0.4436 \pm 0.0266$ & $0.0346 \pm 0.0028$ \\
0.8 & $0.4574 \pm 0.0168$ & $0.0331 \pm 0.0034$ \\
1.0 & $0.4581 \pm 0.0102$ & $0.0333 \pm 0.0018$ \\
\bottomrule
\end{tabular}
\caption{Multi-seed (5 seeds) sweep for EmEL++, Pizza ontology,
GCI1$_\bot$ (disjointness); conventions as in
Table~\ref{app:tab-sweep-b2el-pizza-gci0}.}
\label{app:tab-sweep-emel-pizza-gci1bot}
\end{table}

\begin{table}[h]
\centering
\begin{tabular}{crr}
\toprule
$x$ & FVR $\downarrow$ & MRR $\uparrow$ \\
\midrule
0.0 & $0.8723 \pm 0.0119$ & $0.0921 \pm 0.0074$ \\
0.2 & $0.8866 \pm 0.0128$ & $0.0945 \pm 0.0060$ \\
0.4 & $0.8916 \pm 0.0094$ & $0.0953 \pm 0.0063$ \\
0.6 & $0.8865 \pm 0.0089$ & $0.0941 \pm 0.0085$ \\
0.8 & $0.8891 \pm 0.0030$ & $0.0910 \pm 0.0034$ \\
1.0 & $0.9023 \pm 0.0026$ & $0.0965 \pm 0.0096$ \\
\bottomrule
\end{tabular}
\caption{Multi-seed (5 seeds) sweep for Box2EL, Pizza ontology, GCI2
($C\sqsubseteq\exists r.D$); conventions as in
Table~\ref{app:tab-sweep-b2el-pizza-gci0}.}
\label{app:tab-sweep-b2el-pizza-gci2}
\end{table}

\begin{table}[h]
\centering
\begin{tabular}{crr}
\toprule
$x$ & FVR $\downarrow$ & MRR $\uparrow$ \\
\midrule
0.0 & $0.7406 \pm 0.0038$ & $0.0937 \pm 0.0021$ \\
0.2 & $0.7552 \pm 0.0128$ & $0.0955 \pm 0.0040$ \\
0.4 & $0.7636 \pm 0.0101$ & $0.0974 \pm 0.0028$ \\
0.6 & $0.7732 \pm 0.0105$ & $0.1015 \pm 0.0040$ \\
0.8 & $0.7739 \pm 0.0086$ & $0.1008 \pm 0.0070$ \\
1.0 & $0.7832 \pm 0.0065$ & $0.1022 \pm 0.0053$ \\
\bottomrule
\end{tabular}
\caption{Multi-seed (5 seeds) sweep for ELEmbeddings, Pizza ontology,
GCI2 ($C\sqsubseteq\exists r.D$); conventions as in
Table~\ref{app:tab-sweep-b2el-pizza-gci0}.}
\label{app:tab-sweep-elem-pizza-gci2}
\end{table}

\begin{table}[h]
\centering
\begin{tabular}{crr}
\toprule
$x$ & FVR $\downarrow$ & MRR $\uparrow$ \\
\midrule
0.0 & $0.7269 \pm 0.0041$ & $0.0853 \pm 0.0029$ \\
0.2 & $0.7312 \pm 0.0114$ & $0.0875 \pm 0.0032$ \\
0.4 & $0.7402 \pm 0.0172$ & $0.0911 \pm 0.0019$ \\
0.6 & $0.7526 \pm 0.0088$ & $0.0938 \pm 0.0017$ \\
0.8 & $0.7528 \pm 0.0101$ & $0.0948 \pm 0.0023$ \\
1.0 & $0.7607 \pm 0.0061$ & $0.0976 \pm 0.0022$ \\
\bottomrule
\end{tabular}
\caption{Multi-seed (5 seeds) sweep for EmEL++, Pizza ontology, GCI2
($C\sqsubseteq\exists r.D$); conventions as in
Table~\ref{app:tab-sweep-b2el-pizza-gci0}.}
\label{app:tab-sweep-emel-pizza-gci2}
\end{table}

\begin{table}[h]
\centering
\begin{tabular}{crr}
\toprule
$x$ & FVR $\downarrow$ & MRR $\uparrow$ \\
\midrule
0.0 & $0.4795 \pm 0.0152$ & $0.3941 \pm 0.0327$ \\
0.2 & $0.5259 \pm 0.0326$ & $0.3742 \pm 0.0425$ \\
0.4 & $0.5496 \pm 0.0217$ & $0.3849 \pm 0.0347$ \\
0.6 & $0.5751 \pm 0.0082$ & $0.3772 \pm 0.0402$ \\
0.8 & $0.5815 \pm 0.0257$ & $0.3941 \pm 0.0250$ \\
1.0 & $0.6107 \pm 0.0055$ & $0.3867 \pm 0.0149$ \\
\bottomrule
\end{tabular}
\caption{Multi-seed (5 seeds) sweep for Box2EL, Pizza ontology, GCI3
($\exists r.C\sqsubseteq D$); conventions as in
Table~\ref{app:tab-sweep-b2el-pizza-gci0}.}
\label{app:tab-sweep-b2el-pizza-gci3}
\end{table}

\begin{table}[h]
\centering
\begin{tabular}{crr}
\toprule
$x$ & FVR $\downarrow$ & MRR $\uparrow$ \\
\midrule
0.0 & $0.6566 \pm 0.0077$ & $0.2835 \pm 0.0148$ \\
0.2 & $0.6731 \pm 0.0103$ & $0.2515 \pm 0.0214$ \\
0.4 & $0.6875 \pm 0.0205$ & $0.2410 \pm 0.0203$ \\
0.6 & $0.7211 \pm 0.0350$ & $0.2554 \pm 0.0241$ \\
0.8 & $0.6862 \pm 0.0091$ & $0.2994 \pm 0.0135$ \\
1.0 & $0.7066 \pm 0.0297$ & $0.2263 \pm 0.0400$ \\
\bottomrule
\end{tabular}
\caption{Multi-seed (5 seeds) sweep for ELEmbeddings, Pizza ontology,
GCI3 ($\exists r.C\sqsubseteq D$); conventions as in
Table~\ref{app:tab-sweep-b2el-pizza-gci0}.}
\label{app:tab-sweep-elem-pizza-gci3}
\end{table}

\begin{table}[h]
\centering
\begin{tabular}{crr}
\toprule
$x$ & FVR $\downarrow$ & MRR $\uparrow$ \\
\midrule
0.0 & $0.6340 \pm 0.0050$ & $0.2507 \pm 0.0073$ \\
0.2 & $0.6459 \pm 0.0086$ & $0.2315 \pm 0.0068$ \\
0.4 & $0.6666 \pm 0.0171$ & $0.2144 \pm 0.0203$ \\
0.6 & $0.6897 \pm 0.0138$ & $0.2198 \pm 0.0144$ \\
0.8 & $0.6603 \pm 0.0101$ & $0.2599 \pm 0.0275$ \\
1.0 & $0.6834 \pm 0.0147$ & $0.2128 \pm 0.0101$ \\
\bottomrule
\end{tabular}
\caption{Multi-seed (5 seeds) sweep for EmEL++, Pizza ontology, GCI3
($\exists r.C\sqsubseteq D$); conventions as in
Table~\ref{app:tab-sweep-b2el-pizza-gci0}.}
\label{app:tab-sweep-emel-pizza-gci3}
\end{table}

\end{document}